\documentclass{article} % For LaTeX2e
\usepackage[preprint]{colm2026_conference}

\usepackage{microtype}
\usepackage{hyperref}
\usepackage{url}
\usepackage{booktabs}
\usepackage{amsmath,amssymb}
\usepackage{multirow}
\usepackage{graphicx}
\usepackage[dvipsnames]{xcolor}
\usepackage{wrapfig}
\usepackage{xspace}
\usepackage{listings}
\usepackage{caption}
\usepackage{tcolorbox}
\tcbuselibrary{listings, breakable, skins}
\lstdefinestyle{boxedcode}{
  basicstyle=\ttfamily\small, 
  columns=fullflexible,
  keepspaces=true,
  breaklines=true,
  showstringspaces=false,
  literate={"}{{{\char34}}}1
}

\usepackage{lineno}

\definecolor{darkblue}{rgb}{0, 0, 0.5}
\hypersetup{colorlinks=true, citecolor=darkblue, linkcolor=darkblue, urlcolor=darkblue}

\title{Self-Evolving LLM Memory Extraction \\ Across Heterogeneous Tasks}

\author{Yuqing Yang\textsuperscript{\rm{1}}, 
Tengxiao Liu\textsuperscript{\rm{2}}, 
Wang Bill Zhu\textsuperscript{\rm{1}}, 
Taiwei Shi\textsuperscript{\rm{1}}, 
Linxin Song\textsuperscript{\rm{1}}, 
Robin Jia\textsuperscript{\rm{1}}\thanks{Corresponding author.} \\
\textsuperscript{1}University of Southern California \
\textsuperscript{2}University of California, Santa Barbara \\
\texttt{yyang063@usc.edu} \quad \texttt{robinjia@usc.edu}
}

\newcommand{\methodname}{CluE}
\newcommand{\benchmarkname}{BEHEMOTH\xspace}
\newcommand{\hlfirst}[1]{{\setlength{\fboxsep}{2pt}\colorbox[HTML]{AEC8B5}{#1}}}
\newcommand{\hlsecond}[1]{{\setlength{\fboxsep}{2pt}\colorbox[HTML]{DFE9E2}{#1}}}

\newenvironment{itemize*}%
 {
  \leftmargini=10pt\begin{itemize}%
  }%
 {\end{itemize}}
\newenvironment{enumerate*}%
 {\leftmargini=10pt\begin{enumerate}%
  \setlength{\itemsep}{0pt}%
  \setlength{\parskip}{0pt}}%
 {\end{enumerate}}

\begin{document}

\ifcolmsubmission
\linenumbers
\fi

\maketitle

\begin{abstract}

As LLM-based assistants become persistent and personalized, they must extract and retain useful information from past conversations as memory. However, the types of information worth remembering vary considerably across tasks. We formalize the \textit{heterogeneous memory extraction} task and introduce \textbf{\benchmarkname{}}, a benchmark that repurposes 18 existing datasets spanning personalization, problem-solving, and agentic tasks, using a downstream utility-driven metric for systematic evaluation. Our empirical analysis confirms that no single static extraction prompt dominates across all task categories, and that existing self-evolving prompt optimization frameworks, originally designed for homogeneous distributions, degrade when training tasks are heterogeneous. To address this, we propose \textbf{\methodname{}}, a cluster-based self-evolving strategy that groups training examples into clusters by extraction scenarios, analyzes each cluster independently, and synthesizes cross-cluster insights to update the extraction prompt. Experiments on \benchmarkname{} show that \methodname{} generalizes effectively across heterogeneous tasks ($+$9.04\% relative gain), consistently outperforming prior self-evolving frameworks.

\end{abstract}

\vspace{-4mm}
\section{Introduction}

As large language models (LLMs) become increasingly capable and deeply integrated into daily life, users expect more than isolated, single-session interactions. They want LLMs to remember---to retain personal facts, preferences, and prior experiences across conversations, eliminating the need for repetitive re-specification. Due to long-context constraints \citep{DBLP:journals/tacl/LiuLHPBPL24,DBLP:conf/iclr/WuWYZCY25}, recent work \citep{DBLP:journals/corr/abs-2512-13564,DBLP:journals/tois/ZhangDBMLCZDW25,DBLP:journals/corr/abs-2502-11528} has begun to address this need by equipping LLMs with explicit memory modules. This raises a fundamental question: \textit{how can an LLM determine what information is most worth remembering across diverse interaction scenarios?}

\begin{figure*}[t]
  \centering
  \includegraphics[width=\textwidth]{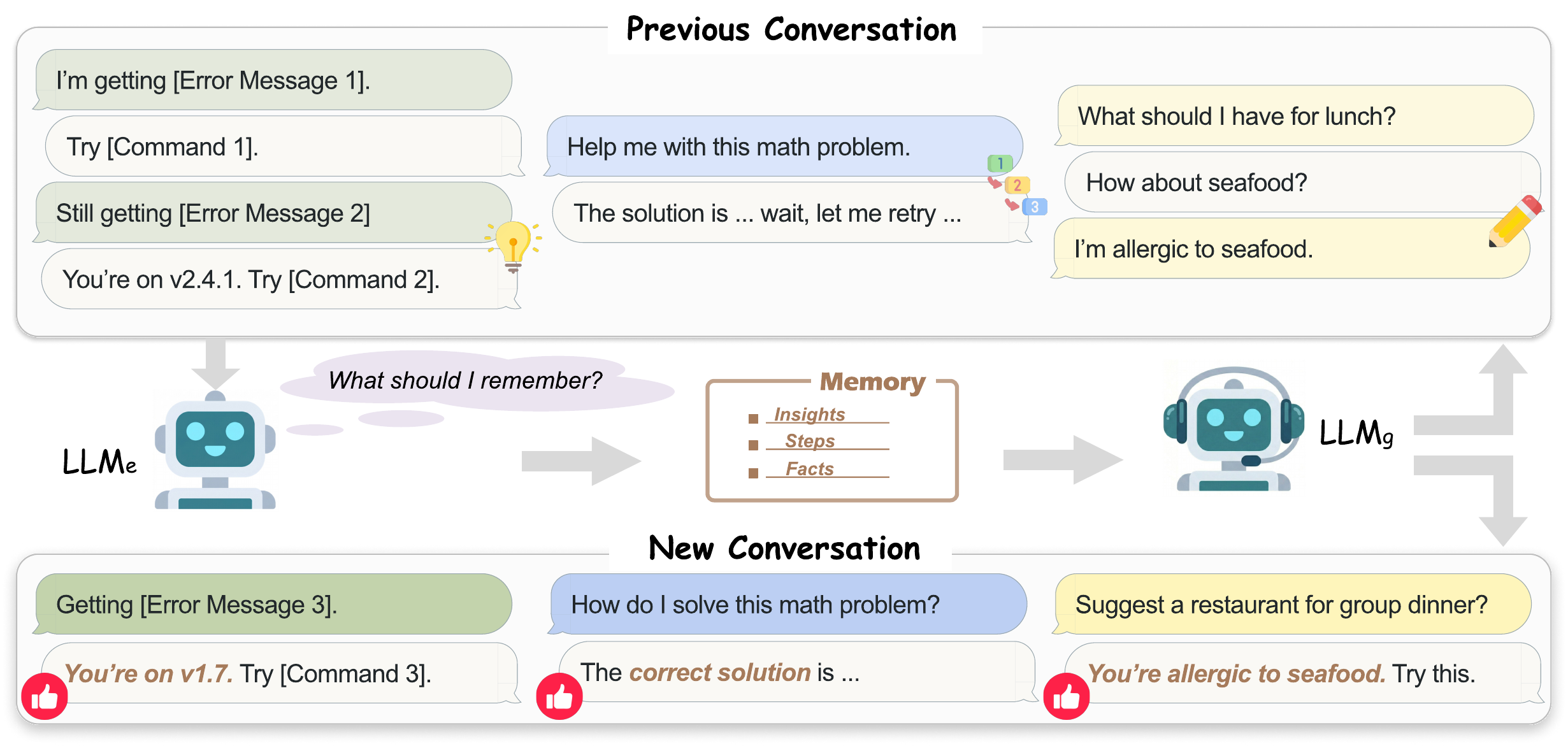}
  % \vspace{-4mm}
  \caption{\small Illustration of heterogeneous memory extraction. A general-purpose assistant $\mathrm{LLM}_g$ encounters diverse previous conversations spanning technical debugging, math problem-solving, and personal preferences, from which an extraction model $\mathrm{LLM}_e$ must produce different types of memory (e.g., reusable insights, solution steps, personal facts). $\mathrm{LLM}_g$ then leverages these memories to improve responses in new conversations. How well these responses address user needs can then serve as a utility-driven signal for evaluating extraction quality.}
  \vspace{-1mm}
  \label{fig:figure1}
  \vspace{-3mm}
\end{figure*}

Existing systems typically answer this question with predefined, static extraction rules tailored to a narrow application context \citep{DBLP:journals/corr/abs-2602-07755}. Conversational agents, for instance, prioritize user preferences and personal facts \citep{DBLP:conf/ecai/ChhikaraKASY25,DBLP:journals/corr/abs-2502-12110}, while agentic systems focus on reusable skills and high-level strategies \citep{DBLP:conf/aaai/Zhao0XLLH24,DBLP:journals/corr/abs-2602-13949}.
However, a general-purpose AI assistant in the real world encounters a far more diverse range of interactions, e.g., from casual conversation to domain-specific problem-solving, and no single, fixed extraction schema can adequately serve this breadth, as illustrated in Figure~\ref{fig:figure1}. Despite this challenge, there is no established benchmark for evaluating the generalizability of memory extraction systems \textbf{across heterogeneous tasks}.

To address this gap, we first formalize single-step memory extraction: given a source conversation, a memory extraction model produces a memory string conditioned on an extraction prompt. We evaluate extraction quality through a downstream, utility-driven metric based on how well the extracted memory helps a generation model answer associated target queries, as illustrated in Figure~\ref{fig:figure1}. Building on this formulation, we propose a benchmark methodology that repurposes existing memory-related datasets and use it to construct \textbf{\benchmarkname{}} (\textbf{B}enchmark for \textbf{E}xtracting \textbf{HE}lpful \textbf{M}emory \textbf{O}n \textbf{T}asks with \textbf{H}eterogeneity), a \textit{heterogeneous} memory extraction benchmark spanning 18 datasets across three task categories (personalization, problem-solving, and agentic tasks). Our evaluation finds that no single static extraction prompt dominates across all task categories, motivating the application of self-evolving frameworks that can automatically discover effective extraction strategies from task feedback.

However, existing self-evolving frameworks including GEPA \citep{DBLP:journals/corr/abs-2507-19457}, ACE \citep{DBLP:journals/corr/abs-2510-04618}, and MemEvolve \citep{DBLP:journals/corr/abs-2512-18746} assume homogeneous task distributions and face new challenges under our heterogeneous distributions. Updating prompts too frequently risks overfitting to recent examples, while updating too infrequently slows adaptation and dilutes feedback signals across dissimilar examples. To navigate this tradeoff, we propose \textbf{\methodname{}} for \textbf{Clu}ster-based \textbf{E}volution. Given a batch of training data, the system first clusters examples into groups based on summarized memory extraction scenarios, performs local analysis within each cluster, and then synthesizes cross-cluster insights to propose updated extraction prompts.
Experiments show that our proposed self-evolving framework consistently outperforms compared methods. All code and data have been publicly available at \url{https://github.com/ayyyq/heterogeneous-memory-extraction}.

In summary, our contributions are as follows:
\vspace{-1mm}
\begin{itemize*}
    \item We formalize the heterogeneous memory extraction task and propose \benchmarkname{}, a benchmark covering 18 datasets for systematic evaluation.
    \item Through evaluation on \benchmarkname{}, we reveal that no single static prompt dominates across all categories, and that existing self-evolving frameworks struggle to generalize across heterogeneous tasks.
    \item We propose \methodname{}, a cluster-based evolution method that achieves stable learning under such heterogeneous distributions, outperforming prior self-evolving frameworks.
\end{itemize*}

\section{Related Work}
\subsection{LLM Memory}

To enable persistent and personalized LLMs, recent work has moved beyond long-context approaches \citep{DBLP:journals/corr/abs-2004-05150}, which suffer from the lost-in-the-middle problem \citep{DBLP:journals/tacl/LiuLHPBPL24}, toward explicitly extracting the information most worth remembering for downstream storage, retrieval and management \citep{DBLP:journals/corr/abs-2512-18746,DBLP:journals/corr/abs-2512-13564}.
Existing systems differ primarily in \emph{what} they extract. One line of work focuses on \textbf{personalized and factual memory}: Mem0 \citep{DBLP:conf/ecai/ChhikaraKASY25} targets preferences, dates, and relationships to construct entity-based graphs; A-Mem \citep{DBLP:journals/corr/abs-2502-12110} extracts keywords and contextual descriptions following Zettelkasten principles. Another line of work targets \textbf{experiential and strategic memory}: ReasoningBank \citep{DBLP:journals/corr/abs-2509-25140} distills successful strategies and failure lessons from agent trajectories; Dynamic Cheatsheet \citep{DBLP:journals/corr/abs-2504-07952} maintains a running summary of reusable problem-solving patterns. Both lines of work rely on fixed, domain-specific extraction rules, limiting their generalizability. From the benchmarking perspective, personalization-focused datasets \citep{DBLP:conf/iclr/WuWYZCY25,DBLP:journals/corr/abs-2512-06688} and problem-solving or agentic datasets \citep{DBLP:journals/corr/abs-2506-07398,DBLP:journals/corr/abs-2504-07952,DBLP:journals/corr/abs-2602-07755} have been used to evaluate memory systems, but no existing benchmark spans all interaction scenarios under a unified protocol. In contrast, we propose a heterogeneous memory extraction benchmark and treat the extraction instruction itself as learnable, motivated by the observation that real-world tasks do not fall neatly into discrete categories and domain-agnostic extraction principles can emerge from heterogeneous task distributions.

\subsection{Self-Evolving Frameworks}

Self-evolving frameworks iteratively refine LLM prompts or strategies from task feedback without manual intervention. General prompt optimization methods such as APE \citep{DBLP:conf/iclr/ZhouMHPPCB23}, OPRO \citep{DBLP:conf/iclr/Yang0LLLZC24}, GEPA \citep{DBLP:journals/corr/abs-2507-19457}, and ACE \citep{DBLP:journals/corr/abs-2510-04618} evolve prompts through various feedback loops, typically developed for single-task benchmarks.
In the memory domain, MemEvolve \citep{DBLP:journals/corr/abs-2512-18746} jointly evolves experiential knowledge and the memory architecture itself through a meta-evolutionary dual loop; ALMA \citep{DBLP:journals/corr/abs-2602-07755} meta-learns entire memory designs as executable code; Evo-Memory \citep{DBLP:journals/corr/abs-2511-20857} evolves the memory content through sequential in-distribution task streams; and MemSkill \citep{DBLP:journals/corr/abs-2602-02474} evolves per-domain memory skill banks using a PPO-trained controller with hard-case clustering. A common assumption across these methods is that the task distribution is homogeneous. Our work contrasts by exploring the potential of self-evolving frameworks on heterogeneous training distributions.

\section{Task Formulation and Dataset Curation}

\subsection{Single-Step Memory Extraction}
\label{sec:task-formulation}

We isolate memory extraction from orthogonal factors such as memory retrieval and management, and formulate a \textbf{single-step memory extraction} task: extract a memory string from one source conversation, and evaluate the extraction quality by applying the memory string as additional context when answering a target query.

Formally, given a \textit{source conversation} $c$ between a user and a generation model $\mathrm{LLM}_g$, the extraction model $\mathrm{LLM}_e$ takes an extraction prompt $P$ as system-level instruction, and the conversation $c$  as input, and produces a memory string $m = \mathrm{LLM}_e(P, c)$. 
To measure extraction quality, we adopt a utility-driven metric rather than LLM-based judges, which suffer from variance and biases \citep{DBLP:conf/iclr/YeWHCZMGG0CC025,DBLP:conf/emnlp/ChenCLJW24}. Given an associated \textit{target query} $q$, the generation model $\mathrm{LLM}_g$ answers $q$ conditioned on the extracted memory: $\hat{y} = \mathrm{LLM}_g(q, m)$. We denote the full interaction for the target query as the target conversation $c_q$.
We use a task-specific \textit{reward function} $R$ to score the response: $r = R(\hat{y}) \in [0, 1]$, where the form of $R$ varies across datasets.

We evaluate memory extraction under two setups: (1) a static prompt $P$ (\S\ref{sec:eval-static}); (2) a self-evolving prompt $P$ (\S\ref{sec:eval-evolving}), optimized iteratively over a training set of triplets $(c, q, R)$.

\subsection{Creating \benchmarkname}

As existing memory benchmarks cover only homogeneous task distributions, we provide a method to construct a \textit{heterogeneous} benchmark that covers diverse tasks under a unified protocol, enabling the first systematic study of memory extraction generalizability.

\begin{wrapfigure}{r}{0.49\linewidth}
    \vspace{-3mm}
    \centering
    \begin{minipage}{\linewidth}
    \centering
    \includegraphics[width=0.95\linewidth]{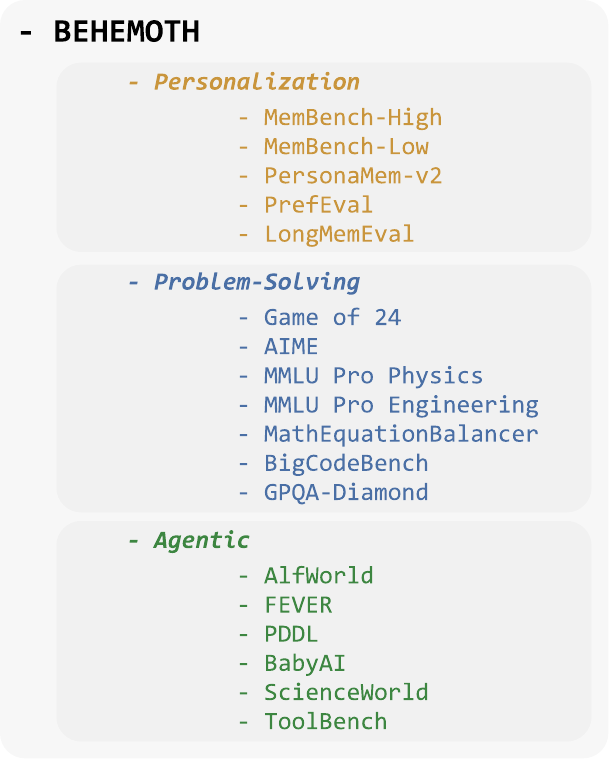}
    \caption{\small Dataset Composition of \benchmarkname{}.}
    \label{fig:benchmark}
    \end{minipage}
    \vspace{-3mm}
\end{wrapfigure}

% \begin{figure}[t]
%     \vspace{-3mm}
%     \centering
%     \includegraphics[width=0.49\linewidth]{figures/benchmark.pdf}
%     \caption{\small Dataset Composition of \benchmarkname{}.}
%     \label{fig:benchmark}
% \end{figure}

Concretely, we repurpose 18 existing datasets into the single-step extraction formulation above and organize them into three categories as presented in Figure~\ref{fig:benchmark}: \textbf{personalization} (5 datasets), involving casual user-assistant conversations; \textbf{problem-solving} (7 datasets), involving reasoning problems; and \textbf{agentic} (6 datasets), involving multi-turn action-feedback trajectories.
We use Qwen3-32B~\citep{DBLP:journals/corr/abs-2505-09388} as $\mathrm{LLM}_g$ to complete source conversations for the 18 datasets, and refer to the resulting benchmark as \textbf{\benchmarkname}, where each instance is a triplet $(c, q, R)$. 
The curation pipeline is general and readily extensible, as it is parameterized by the choice of datasets and $\mathrm{LLM}_g$. 

To evaluate whether a self-evolving prompt generalizes beyond the training distribution, we hold out one dataset per category, i.e., LongMemEval~\citep{DBLP:conf/iclr/WuWYZCY25}, GPQA-Diamond~\citep{DBLP:journals/corr/abs-2311-12022}, and ToolBench~\citep{DBLP:conf/acl/GuoCWLQLL0L24}, as out-of-distribution test sets. 
From the remaining in-distribution datasets, we randomly sample 20 examples from each (50 for AIME, which aggregates multiple competition years) to form the training set (330 in total, randomly shuffled) and use the rest as in-distribution test sets.
See Appendix~\ref{app:benchmark} for details of benchmark construction and statistics.

Importantly, the three categories do not determine what type of memory should be extracted from each example. Within the same category, a successful agent trajectory calls for reusable strategies while a failed one calls for pitfalls to avoid; across categories, extracting domain facts from a personalization dialogue is analogous to extracting domain knowledge from a problem-solving session. Since the memory extraction system receives only the raw conversation without any category label, it must \textit{discover} the appropriate extraction behavior from the data itself.

\subsection{Evaluation Metrics}

Let $J_{D_k}(P) = \frac{1}{|D_k|} \sum_{D_k} r$ denote the average reward on dataset $D_k$. Although the reward depends on the underlying models, we fix the model configuration across all experiments and omit it from the notation, writing the metrics as functions of the extraction prompt $P$ alone. We use the following two metrics to evaluate the generalizability of $P$ across $N$ datasets for both static and evolving setups:
\[
    \textbf{Macro Accuracy}(P) = \frac{1}{N}\sum_{k=1}^{N} J_{D_k}(P); \qquad 
    \textbf{Relative Gain}(P) = \left(\prod_{k=1}^{N} \frac{J_{D_k}(P)}{J_{D_k}(P_{\text{base}})}\right)^{1/N} - 1.
\]
\textbf{Macro accuracy} weights all datasets equally regardless of size, measuring absolute cross-domain performance. \textbf{Relative gain} measures the geometric mean of per-dataset improvement ratios over a baseline prompt $P_{\text{base}}$. Dividing by the baseline normalizes away absolute score differences, and taking the geometric (rather than arithmetic) mean prevents a few outlier ratios from dominating the aggregate.

\section{Evaluating Static Memory Extraction}
\label{sec:eval-static}

We first evaluate static memory extraction prompts, either hand-crafted for specific tasks or summarized from existing work. In Table~\ref{tab:evaluation-static}, \textit{No Memory} provides no memory to the generation model, serving as a baseline, and \textit{Simple} uses a minimal prompt that asks the model to extract useful information without specifying any taxonomy. The remaining four are organized into two groups: \textbf{category-specific} prompts, including \textit{Mem0}~\citep{DBLP:conf/ecai/ChhikaraKASY25}, which targets user preferences and personal facts, and \textit{ReasoningBank}~\citep{DBLP:journals/corr/abs-2509-25140}, which focuses on extracting successful strategies and failure lessons from trajectories; and \textbf{taxonomy-based} prompts, including \textit{OpenMemory}\footnote{\url{https://openmemory.cavira.app/}}, which defines a five-class memory taxonomy, and \textit{Survey}~\citep{DBLP:journals/corr/abs-2512-13564}, which classifies memories into two broad types. The full prompts are provided in Appendix~\ref{app:static-prompts}. Unless otherwise stated, we use Qwen3-32B as both the generation model and the extraction model throughout all experiments.

\begin{table}[t]
    \small
    \centering
    \resizebox{\linewidth}{!}{
    \begin{tabular}{lcccccccc}
        \toprule
         & \multicolumn{2}{c}{\textbf{Personalization}} & \multicolumn{2}{c}{\textbf{Problem-Solving}} & \multicolumn{2}{c}{\textbf{Agentic}} & \multicolumn{2}{c}{\textbf{Overall}} \\
         \cmidrule(lr){2-3}\cmidrule(lr){4-5}\cmidrule(lr){6-7}\cmidrule(lr){8-9}
         & \textbf{MA} & \textbf{RG} & \textbf{MA} & \textbf{RG} & \textbf{MA} & \textbf{RG} & \textbf{MA} & \textbf{RG} \\
         \midrule
         \textbf{No Memory} & 34.15 & -- & 46.52 & -- & 30.36 & -- & 37.84 & -- \\
         \textbf{Simple} & 58.76 & 0 & 48.76 & 0 & 32.46 & 0 & 46.00 & 0 \\
         \textbf{Mem0} & \hlfirst{73.31} & \textbf{$+$29.96} & 45.97 & $-$3.70 & 31.62 & $+$0.44 & \hlfirst{48.48} & \textbf{$+$5.79} \\
         \textbf{ReasoningBank} & 50.67 & $-$10.78 & \hlfirst{50.72} & \textbf{$+$6.90} & 32.12 & $+$2.49 & 44.51 & $+$0.45 \\
         \textbf{OpenMemory} & 57.39 & $-$1.04 & \hlsecond{50.47} & $+$5.59 & \hlfirst{34.93} & \textbf{$+$8.57} & 47.14 & $+$4.74 \\
         \textbf{Survey} & \hlsecond{62.05} & $+$7.35 & 49.86 & $+$4.80 & \hlsecond{33.58} & $+$2.89 & \hlsecond{47.69} & $+$4.83 \\
        \bottomrule
    \end{tabular}%
    }
    \caption{\small Static prompt evaluation on in-distribution test sets. \textbf{MA}: Macro Accuracy (\%); \textbf{RG}: Relative Gain (\%) over \textit{Simple}. \hlfirst{Best} and \hlsecond{second best} MA are highlighted; best RG in \textbf{bold}. \textbf{Overall} aggregates across all datasets.}
    \label{tab:evaluation-static}
    \vspace{-4mm}
\end{table}

We have the following observations from Table~\ref{tab:evaluation-static}: \textbf{(1) Using memory consistently improves over no-memory baselines.} Even the minimal \textit{Simple} prompt raises overall macro accuracy from 37.84\% to 46.00\%, with gains across all three categories, confirming the value of memory extraction. \textbf{(2) No single static memory extraction prompt dominates.} Category-specific memory extraction prompts (i.e., \textit{Mem0} and \textit{ReasoningBank}) excel in their target domains but underperform elsewhere, while taxonomy-based memory extraction prompts (i.e., \textit{Survey} and \textit{OpenMemory}) offer more balanced but moderate improvements. \textbf{(3) A more detailed taxonomy does not guarantee better extraction.} \textit{OpenMemory}'s five-class taxonomy does not outperform \textit{Survey}'s simpler two-class design, indicating that extraction quality is not solely determined by the granularity of the prompt but also by how well the extraction model can interpret and follow it.

\vspace{-1mm}
\section{Evaluating Evolving Memory Extraction}
\label{sec:eval-evolving}
\vspace{-1mm}

The observations above, namely the trade-off between specialization and coverage and the difficulty of manually designing a universally effective prompt, motivate automatically discovering effective extraction strategies from task feedback. We propose \textbf{\methodname{}}, a self-evolving framework that enables stable learning under heterogeneous training distributions, and compare it with three popular self-evolving methods.

\subsection{Baselines}

We describe the optimization procedure of each baseline below; implementation details are provided in Appendix~\ref{app:self-evolve}.

\vspace{-1mm}
\paragraph{GEPA.} GEPA~\citep{DBLP:journals/corr/abs-2507-19457} employs a reflective Proposer that takes the current prompt together with training logs (source conversation $c$, extracted memory $m$, target conversation $c_q$, and reward $r$) and proposes a refined prompt. Because both the prompt and the logs must fit in the LLM context, each update is limited to a small batch of examples.

\vspace{-1mm}
\paragraph{ACE.} ACE~\citep{DBLP:journals/corr/abs-2510-04618} consists of a Reflector and a Curator. The Reflector analyzes each extraction attempt to identify reasons for success or failure and tags rules in the prompt as helpful or harmful. The Curator then applies atomic operations (add, update, or delete) to the prompt based on the accumulated helpful/harmful counts.

\vspace{-1mm}
\paragraph{MemEvolve.} MemEvolve~\citep{DBLP:journals/corr/abs-2512-18746} consists of an Analyzer and a Proposer. The Analyzer is implemented as a tool-augmented agent that can retrieve and inspect individual training logs by ID, allowing it to process a large batch of data. The Proposer then refines the prompt based on the Analyzer's findings.

\subsection{CluE: Cluster-based Evolution}

\begin{figure*}[t]
  \centering
  \includegraphics[width=\textwidth]{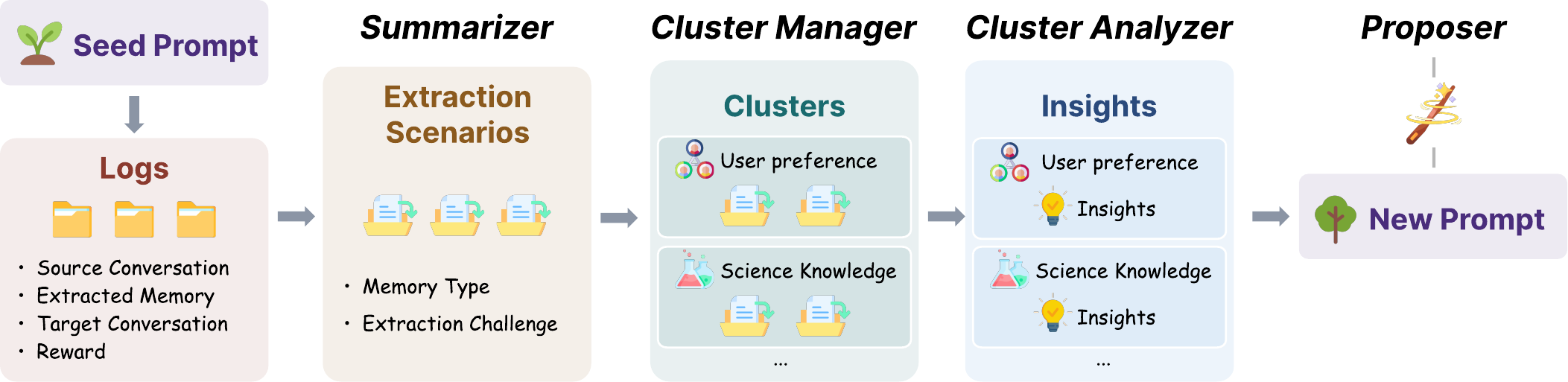}
  \caption{\small Overview of our method \methodname{} in one evolution round.}
  \label{fig:method}
  % \vspace{-4mm}
\end{figure*}

The above frameworks were designed for homogeneous distributions where all examples share similar features. Under heterogeneous tasks, they face a dilemma: updating too frequently causes instability, as successive batches of dissimilar examples pull the prompt in conflicting directions; updating too infrequently causes dilution, as feedback from diverse tasks is averaged together, washing out fine-grained insights.

To address this, we follow MemEvolve's analyzer-proposer architecture and propose \textbf{\methodname{}} (\textbf{Clu}ster-based \textbf{E}volution), which introduces \textbf{cluster-based updates}: training examples are grouped into clusters that share similar \textit{memory extraction scenarios}, and analysis is performed per cluster rather than over the entire batch.

The system operates in rounds; each round evaluates the current prompt on a training batch to produce logs, then updates the prompt in the following four steps, as illustrated in Figure~\ref{fig:method}. The prompts for each step are provided in Appendix~\ref{app:clue-prompt}.

\paragraph{Summarization.} For each example in the batch, a Summarizer reads its log and generates a concise summary describing its \textit{extraction scenario}, i.e., what type of information needs to be extracted (e.g., procedural steps, user preferences, causal reasoning chains) and what makes extraction challenging (e.g., long context, implicit information, multi-turn reasoning). These summaries abstract away surface-level details (dataset name, specific content) and focus on properties relevant to memory extraction.

\paragraph{Clustering.} A Cluster Manager takes all summaries from the current batch together with existing cluster labels and descriptions, and assigns each example to a cluster by extraction scenario, regardless of its original dataset or category. For instance, ``extracting procedural knowledge from lengthy dialogues'' may group together examples from both agentic trajectories and step-by-step math solutions. The Cluster Manager may create new clusters, merge existing ones, or split them as the distribution of examples evolves across rounds.

\paragraph{Cluster-based Analysis.} For each cluster, a Cluster Analyzer retrieves and inspects the training logs of that cluster's examples. It identifies success patterns (what makes extracted memory useful for this scenario), failure patterns (what is missing or ineffective), and produces targeted recommendations. Each cluster is analyzed independently, so recommendations for one scenario do not interfere with another.

\paragraph{Cross-cluster Proposal.} All per-cluster analyses are fed into a Proposer that generates an improved memory extraction prompt. It identifies general principles shared across clusters, organizes cluster-specific insights into structured guidelines, and resolves conflicting recommendations by scoping them to relevant memory categories. The resulting prompt captures both cross-domain principles and scenario-specific guidance, producing targeted modifications rather than full rewrites.

\subsection{Results}
\label{sec:eval-simple}

\begin{table}[t]
    \small
    \centering
    \resizebox{\linewidth}{!}{
    \begin{tabular}{lcccccccc}
        \toprule
         & \multicolumn{2}{c}{\textbf{Personalization}} & \multicolumn{2}{c}{\textbf{Problem-Solving}} & \multicolumn{2}{c}{\textbf{Agentic}} & \multicolumn{2}{c}{\textbf{Overall}} \\
         \cmidrule(lr){2-3}\cmidrule(lr){4-5}\cmidrule(lr){6-7}\cmidrule(lr){8-9}
         & \textbf{MA} & \textbf{RG} & \textbf{MA} & \textbf{RG} & \textbf{MA} & \textbf{RG} & \textbf{MA} & \textbf{RG} \\
         \midrule
         \textbf{No Memory} & 34.15 & -- & 46.52 & -- & 30.36 & -- & 37.84 & -- \\
         \textbf{Simple} & 58.76 & 0 & 48.76 & 0 & 32.46 & 0 & 46.00 & 0 \\
         \midrule
         \textbf{GEPA} & 56.59 & $-$2.16 & 50.05 & $+$2.86 & \hlfirst{37.23} & \textbf{$+$14.08} & \hlsecond{47.52} & $+$5.06 \\
         \textbf{ACE} & 53.73 & $-$4.38 & \hlsecond{51.04} & $+$8.04 & 33.49 & $+$4.93 & 45.91 & $+$3.56 \\
         \textbf{MemEvolve} & \hlsecond{63.67} & $+$10.76 & 49.49 & $+$1.17 & 30.92 & $-$3.25 & 47.08 & $+$2.11 \\
         \textbf{\methodname{}} & \hlfirst{65.72} & \textbf{$+$12.34} & \hlfirst{51.85} & \textbf{$+$8.39} & \hlsecond{35.24} & $+$7.22 & \hlfirst{50.01} & \textbf{$+$9.04} \\
        \bottomrule
    \end{tabular}}
    \caption{\small Evaluation of self-evolving frameworks starting from the \textit{Simple} prompt on in-distribution test sets. \textbf{RG} denotes the Relative Gain (\%) over \textit{Simple}.}
    \label{tab:evaluation-simple}
    % \vspace{-2mm}
\end{table}

We evaluate self-evolving frameworks starting from the \textit{Simple} prompt and report in-distribution and out-of-distribution results separately.

\paragraph{Existing frameworks improve unevenly across categories.} As shown in Table~\ref{tab:evaluation-simple}, all self-evolving frameworks achieve positive overall relative gains over the \textit{Simple} baseline, but each exhibits a characteristic bias: gains on some categories come at the cost of regressions on others. GEPA achieves the strongest agentic improvement ($+$14.08\%) yet regresses on personalization ($-$2.16\%). MemEvolve excels at personalization (+10.76\%) but drops on agentic tasks ($-$3.25\%). This pattern demonstrates that under heterogeneous distributions, these frameworks trade off performance across tasks rather than improving them jointly.

\paragraph{\methodname{} achieves the best overall performance among self-evolving methods.} Our method yields the highest overall relative gain ($+$9.04\%), with consistent improvements across all three categories ($+$12.34\% personalization, $+$8.39\% problem-solving, $+$7.22\% agentic). Moreover, when switching the extraction model from Qwen3-32B to Gemini-3-Flash \citep{doshi2025gemini3flash}, \methodname{} still outperforms all baseline self-evolving frameworks (Appendix~\ref{app:eval-gemini}).

\begin{wraptable}{r}{0.5\linewidth}
    % \small
    \centering
    \vspace{-2mm}
    \resizebox{\linewidth}{!}{
    \begin{tabular}{lccc}
        \toprule
         & \textbf{LME} & \textbf{GPQA-D} & \textbf{ToolBench} \\
         \midrule
         \textbf{No Memory} & 20.45 & 45.62$_{\pm 3.12}$ & 23.33 \\
         \textbf{Simple} & 46.02 & 47.14$_{\pm 4.15}$ & 25.30 \\
         \midrule
         \textbf{GEPA} & 35.06 & \hlfirst{50.00}$_{\pm 0.82}$ & 24.09 \\
         \textbf{ACE} & 29.71 & 46.13$_{\pm 0.86}$ & \hlfirst{26.82} \\
         \textbf{MemEvolve} & \hlsecond{56.82} & 47.98$_{\pm 1.09}$ & \hlsecond{26.67} \\
         \textbf{\methodname{}} & \hlfirst{63.07} & \hlsecond{48.48}$_{\pm 1.80}$ & \hlsecond{26.67} \\
        \bottomrule
    \end{tabular}}
    \caption{\small Evaluation on the out-of-distribution test sets of \benchmarkname{}.}
    \label{tab:evaluation-simple-ood}
    \vspace{-1em}
\end{wraptable}

\paragraph{Generalization to out-of-distribution tasks.} The diversity of the in-distribution test sets already provides evidence of generalization. To further validate this, we evaluate the evolved prompts on held-out datasets not used during evolution (Table~\ref{tab:evaluation-simple-ood}). Our method achieves the strongest result on LongMemEval (63.07 vs.\ 56.82 for the second best) and does not drop below the \textit{Simple} prompt on GPQA-Diamond or ToolBench, whereas GEPA degrades on ToolBench and ACE degrades on GPQA-Diamond, confirming that our method yields broadly applicable extraction prompts rather than ones overfit to specific datasets.

\section{Further Analysis}

\subsection{Evolution From a Stronger Seed}

\begin{table}[ht]
    \small
    \centering
    \resizebox{\linewidth}{!}{
    \begin{tabular}{lcccccccc}
        \toprule
         & \multicolumn{2}{c}{\textbf{Personalization}} & \multicolumn{2}{c}{\textbf{Problem-Solving}} & \multicolumn{2}{c}{\textbf{Agentic}} & \multicolumn{2}{c}{\textbf{Overall}} \\
         \cmidrule(lr){2-3}\cmidrule(lr){4-5}\cmidrule(lr){6-7}\cmidrule(lr){8-9}
         & \textbf{MA} & \textbf{RG} & \textbf{MA} & \textbf{RG} & \textbf{MA} & \textbf{RG} & \textbf{MA} & \textbf{RG} \\
         \midrule
         \textbf{No Memory} & 34.15 & -- & 46.52 & -- & 30.36 & -- & 37.84 & -- \\
         \textbf{Survey} & 62.05 & 0 & 49.86 & 0 & 33.58 & 0 & 47.69 & 0 \\
         \midrule
         \textbf{GEPA$^\dagger$} & 62.05 & 0 & 49.86 & 0 & 33.58 & 0 & 47.69 & 0 \\
         \textbf{ACE} & 54.80 & $-$10.48 & 50.06 & $+$0.75 & \hlsecond{34.41} & $+$3.69 & 46.11 & $-$1.44 \\
         \textbf{MemEvolve} & \hlsecond{62.88} & $-$0.31 & \hlsecond{51.22} & $+$2.51 & 31.34 & $-$4.85 & \hlsecond{47.70} & $-$0.74 \\
         \textbf{\methodname{}} & \hlfirst{70.67} & \textbf{$+$13.62} & \hlfirst{51.63} & \textbf{$+$2.68} & \hlfirst{34.69} & \textbf{$+$5.79} & \hlfirst{51.06} & \textbf{$+$6.54} \\
        \bottomrule
    \end{tabular}%
    }
    \caption{\small Evaluation of self-evolving frameworks starting from the \textit{Survey} prompt on in-distribution test sets. \textbf{RG} denotes the Relative Gain (\%) over \textit{Survey}. $\dagger$ GEPA did not find a better prompt.}
    \label{tab:evaluation-seed}
    % \vspace{-4mm}
\end{table}

In \S\ref{sec:eval-simple}, evolution starts from a minimal seed (\textit{Simple}). Here we test whether each method can leverage a stronger starting point by initializing from the \textit{Survey} prompt. A stronger seed leaves less room for improvement but more room for damage: undirected updates are more likely to overwrite useful guidelines already present in the seed. As shown in Table~\ref{tab:evaluation-seed}, GEPA fails to find any improvement and returns the seed unchanged. ACE and MemEvolve both produce negative overall relative gains ($-$1.44\% and $-$0.74\%): while they achieve marginal improvements on some categories, these are offset by notable regressions elsewhere (e.g., ACE drops $-$10.48\% on personalization, MemEvolve drops $-$4.85\% on agentic), indicating that their updates inadvertently damage existing strengths of the seed. In contrast, \methodname{} achieves $+$6.54\% overall relative gain with improvements across all three categories, demonstrating that cluster-based analysis can preserve the seed's strengths while still discovering room for improvement.

\subsection{From Single-Step to Continual Memory Extraction}
\label{sec:continual}

Our main setup adopts a single-step protocol for efficient and stable evaluation. In practice, however, memory systems operate \emph{continually}: conversations arrive sequentially, previously extracted memories are retrieved and injected into future conversations, and new extractions must operate on this augmented context. In this section, we examine whether the advantages observed in the single-step setting carry over to the continual setting.

Following \citet{DBLP:journals/corr/abs-2509-25140}, we adopt a simple continual pipeline---embedding-based top-$k$ ($k=1$) retrieval plus concatenation-based consolidation---to isolate the effect of extraction prompt quality from retrieval and consolidation design. We select two tasks on which \methodname{} outperforms MemEvolve in the single-step evaluation: Game of 24 (problem-solving) and AlfWorld (agentic), and focus the analysis on whether single-step gains transfer.

\begin{table}[ht]
    \centering
    % \small
    % \vspace{-1mm}
    \begin{tabular}{lcccc}
        \toprule
         & \multicolumn{2}{c}{\textbf{Game of 24}} & \multicolumn{2}{c}{\textbf{AlfWorld}} \\
         \cmidrule(lr){2-3}\cmidrule(lr){4-5}
         & \textbf{Single-Step} & \textbf{Continual} & \textbf{Single-Step} & \textbf{Continual} \\
         \midrule
         \textbf{No Memory} & $26.67_{\pm 3.86}$ & $35.42_{\pm 5.03}$ & $30.41_{\pm 2.71}$ & $63.74_{\pm 0.41}$ \\
         \textbf{Simple} & $\hlsecond{27.92}_{\pm 3.86}$ & $29.58_{\pm 3.12}$ & $\hlsecond{47.66}_{\pm 2.19}$ & $\hlsecond{64.91}_{\pm 5.01}$ \\
         \textbf{MemEvolve} & $27.50_{\pm 2.70}$ & $\hlsecond{43.33}_{\pm 3.58}$ & $40.64_{\pm 0.41}$ & $62.57_{\pm 2.30}$ \\
         \textbf{\methodname{}} & $\hlfirst{37.08}_{\pm 2.12}$ & $\hlfirst{50.83}_{\pm 7.52}$ & $\hlfirst{55.85}_{\pm 3.23}$ & $\hlfirst{67.25}_{\pm 2.98}$ \\
        \bottomrule
    \end{tabular}%
    \caption{Single-step vs.\ continual memory extraction on Game of 24 and AlfWorld. The \textit{No Memory} baselines differ between settings because the two settings evaluate on different example pools; see Appendix~\ref{app:continual-details} for details.}
    \label{tab:continual}
\end{table}

As shown in Table~\ref{tab:continual}, \methodname{} maintains its advantage over MemEvolve in both tasks under the continual setting (50.83 vs.\ 43.33 on Game of 24; 67.25 vs.\ 62.57 on AlfWorld), confirming that the single-step gains transfer when memories are accumulated and retrieved over time. Moreover, the \textit{Simple} prompt, which lacks detailed extraction guidelines, falls below the \textit{No Memory} baseline on average (Game of 24) or within its variance (AlfWorld), whereas \methodname{} outperforms \textit{No Memory} more stably on both tasks. This highlights that well-evolved extraction guidelines are especially important in the continual setting, where low-quality memories accumulate and compound over time.

\subsection{Qualitative Analysis of Evolved Prompts}

\begin{figure*}[t]
  \centering
  \includegraphics[width=\textwidth]{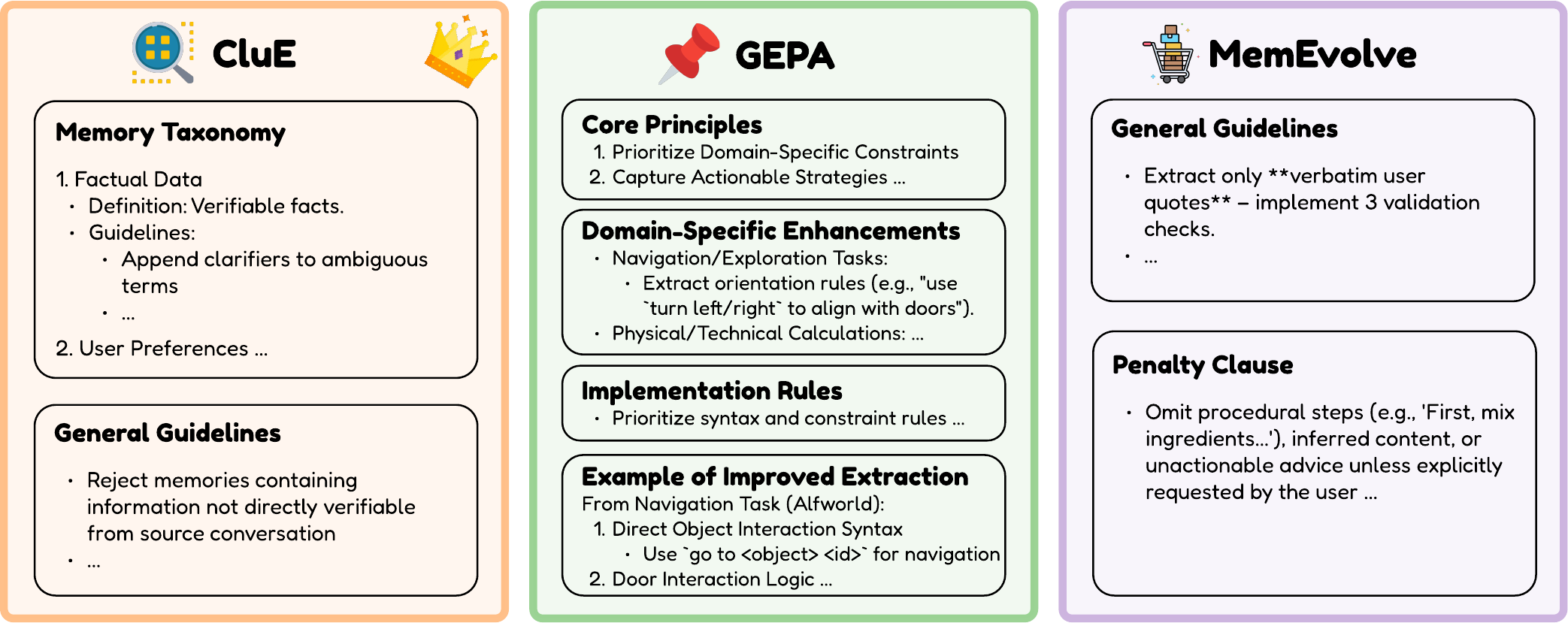}
  \caption{\small Structural comparison of the best prompts evolved by each framework from the \textit{Simple} seed. Common sections (task description, input/output format) are omitted; only the distinctive components are shown. \methodname{} produces a structured memory taxonomy with per-category definitions and guidelines alongside domain-agnostic general guidelines. GEPA embeds a large amount of domain-specific content (e.g., AlfWorld examples) directly into its rules. MemEvolve lacks a memory taxonomy and instead introduces a penalty clause that prohibits specific extraction patterns.}
  \label{fig:prompt}
  \vspace{-4mm}
\end{figure*}

\label{sec:qualitative}

To understand \emph{why} the quantitative gaps arise, we inspect the best prompts produced by each framework after evolution from the \textit{Simple} prompt. ACE produces a structurally similar prompt to \methodname{} but is substantially more verbose, as its atomic updates predominantly add rules and rarely delete them. The resulting prompt (1,403 tokens) is considerably longer than those of GEPA (1,243), \methodname{} (936), and MemEvolve (539), which may weaken the extraction model's instruction-following. We illustrate the key structural differences for the remaining frameworks in Figure~\ref{fig:prompt}.

According to Figure~\ref{fig:prompt}, GEPA's fixed context-length budget limits each update to a small batch, biasing the prompt toward the most recent examples (in this case, AlfWorld). This domain-specific bias explains its strong agentic gains but regression on other categories.
MemEvolve analyzes large batches spanning diverse tasks without category-aware grouping, causing heterogeneous feedback signals to cancel out. As a result, its evolved prompt lacks a specific memory taxonomy and retains only known failure modes.
Although \methodname{} uses the same batch size as MemEvolve, cluster-based analysis yields a qualitatively different outcome: a complete memory taxonomy with category-specific guidelines as well as domain-agnostic principles, explaining the robust generalization across heterogeneous tasks and out-of-distribution datasets. The fully evolved prompt of \methodname{} is shown in Figure~\ref{fig:prompt_evolved}.
We further examine how the clusters evolve during evolution and how they are transformed into the final prompt in Appendix~\ref{app:cluster-evolution}.

\section{Discussion and Conclusions}

In this work, we focus on LLM memory extraction under heterogeneous tasks. We formalize single-step memory extraction, introduce a general benchmark methodology instantiated as \benchmarkname{}, and use it to expose the limitations of both static extraction prompts and existing self-evolving frameworks. To overcome these limitations, we propose \methodname{}, a cluster-based evolution framework that enables stable learning under heterogeneous training distributions, and demonstrate its effectiveness through extensive experiments.

We believe both artifacts of this work may find use beyond the scope studied here. \benchmarkname{} can serve as a testbed for evaluating a wide range of memory extraction approaches---not only self-evolving frameworks but also routing-based, skill-based, and other emerging paradigms---to explore what strategies best handle heterogeneous memory extraction. The self-evolving framework \methodname{} may also extend beyond memory extraction to any setting where a single agent must handle heterogeneous demands, such as serving multiple users with distinct habits and communication styles, or supporting the same user across projects that span different domains and workflows.

Several directions remain open. First, although \benchmarkname{} covers 18 datasets, real-world deployments present more complex scenarios with greater task diversity and longer interaction histories; constructing more challenging and realistic benchmarks is an important next step. Second, this work focuses on memory extraction as the first stage of the memory lifecycle; achieving generalizable performance across the full lifecycle, including storage, retrieval, and management, remains an open challenge.

% \section*{Author Contributions}
% If you'd like to, you may include  a section for author contributions as is done
% in many journals. This is optional and at the discretion of the authors.

\section*{Acknowledgments}
%Use unnumbered first level headings for the acknowledgments. All
%acknowledgments, including those to funding agencies, go at the end of the paper.

We thank Ameya Godbole for thoughtful discussions and valuable feedback on this work.
This work was supported in part by the National Science Foundation under Grant No. IIS-2403436. Any opinions, findings, and conclusions or recommendations expressed in this material are those of the author(s) and do not necessarily reflect the views of the National Science Foundation.

% \section*{Ethics Statement}
% Authors can add an optional ethics statement to the paper. 
% For papers that touch on ethical issues, this section will be evaluated as part of the review process. The ethics statement should come at the end of the paper. It does not count toward the page limit, but should not be more than 1 page. 

\bibliography{colm2026_conference}
\bibliographystyle{colm2026_conference}

\newpage
\appendix

% \section*{Disclosure of LLM Usage}

% As described throughout the paper, LLMs (Qwen3-32B and Gemini-3-Flash) are central to our research as both the extraction and generation models under study. We also use LLM-as-judge evaluation for LongMemEval \citep{DBLP:conf/iclr/WuWYZCY25} and ToolBench \citep{DBLP:conf/acl/GuoCWLQLL0L24}. Beyond their role as research 
% subjects and evaluation tools, we did not use LLMs to originate research ideas, write original content in the paper, or generate data or plots.

\section{Experimental Details}

\subsection{\benchmarkname{} Curation}
\label{app:benchmark}

Following the task formulation in Section~\ref{sec:task-formulation}, each example in \benchmarkname{} consists of a source conversation $c$, a target query $q$, and a reward function $R$. For personalization datasets, which already contain multi-turn dialogues, we directly use the provided source conversations and target queries. Problem-solving and agentic datasets, however, typically supply only task instances rather than full conversations; we therefore prompt $\mathrm{LLM}_g$ to solve each task instance and use the resulting trajectory as the source conversation, and following the retrieval setup of prior memory work (Table~\ref{tab:benchmark}), retrieve a semantically similar query as the target query. As a result, each example's source conversation originates from a distinct query, but multiple examples may share the same target query.

The dataset statistics are summarized in Table~\ref{tab:benchmark}. For each dataset, we randomly sample 20 examples as the in-distribution training set (50 for AIME, which aggregates multiple competition years) and use the remainder (capped at 200) as the test set. During training (evolution), examples from all datasets are randomly shuffled rather than grouped by dataset, so each batch may contain examples from multiple datasets. For test sets with fewer than 200 examples, we repeat the full evaluation pipeline (memory extraction by $\mathrm{LLM}_e$ and target query answering by $\mathrm{LLM}_g$) three times and report the average for all experiments.

\begin{table}[ht]
    \small
    \centering
    \resizebox{\linewidth}{!}{
    \begin{tabular}{lcccll}
        \toprule
        \textbf{Dataset} & \textbf{\# Train} & \textbf{\# Test} & \textbf{$R$} & \textbf{Memory Work} \\
        \midrule
        \multicolumn{5}{c}{\textbf{\textit{Personalization}}} \\
        \midrule
        MemBench-High~\citep{DBLP:conf/acl/Tan000DD25} & 20 & 200 & Acc. & MemBench~\citep{DBLP:conf/acl/Tan000DD25} \\
        MemBench-Low~\citep{DBLP:conf/acl/Tan000DD25} & 20 & 80 & Acc. & MemBench~\citep{DBLP:conf/acl/Tan000DD25} \\
        PersonaMem-v2~\citep{DBLP:journals/corr/abs-2512-06688} & 20 & 272 & Acc. & PersonaMem-v2~\citep{DBLP:journals/corr/abs-2512-06688} \\
        PrefEval~\citep{DBLP:conf/iclr/Zhao00HL25} & 20 & 200 & Acc. & PrefEval~\citep{DBLP:conf/iclr/Zhao00HL25} \\
        LongMemEval~\citep{DBLP:conf/iclr/WuWYZCY25} & -- & 176 & LLM Judge & LongMemEval~\citep{DBLP:conf/iclr/WuWYZCY25} \\
        % \midrule
        \midrule
        \multicolumn{5}{c}{\textbf{\textit{Problem-Solving}}} \\
        \midrule
        Game of 24~\citep{DBLP:journals/corr/abs-2401-12954} & 20 & 80 & Rule & DC~\citep{DBLP:journals/corr/abs-2504-07952} \\
        AIME & 50 & 143 & EM & DC~\citep{DBLP:journals/corr/abs-2504-07952} \\
        MMLU Pro Physics~\citep{DBLP:conf/nips/WangMZNCGRAHJLK24} & 20 & 200 & Acc. & DC~\citep{DBLP:journals/corr/abs-2504-07952} \\
        MMLU Pro Engineering~\citep{DBLP:conf/nips/WangMZNCGRAHJLK24} & 20 & 200 & Acc. & DC~\citep{DBLP:journals/corr/abs-2504-07952} \\
        MathEquationBalancer & 20 & 200 & Rule & DC~\citep{DBLP:journals/corr/abs-2504-07952} \\
        BigCodeBench~\citep{DBLP:conf/iclr/ZhuoVCH0WYZHPB025} & 20 & 200 & Pass@1 & MemGen~\citep{DBLP:journals/corr/abs-2509-24704} \\
        GPQA-Diamond~\citep{DBLP:journals/corr/abs-2311-12022} & -- & 198 & Acc. & DC~\citep{DBLP:journals/corr/abs-2504-07952} \\
        % \midrule
        \midrule
        \multicolumn{5}{c}{\textbf{\textit{Agentic}}} \\
        \midrule
        AlfWorld~\citep{DBLP:conf/iclr/ShridharYCBTH21} & 20 & 114 & SR & G-Memory~\citep{DBLP:journals/corr/abs-2506-07398} \\
        FEVER~\citep{DBLP:conf/naacl/ThorneVCM18} & 20 & 200 & SR & G-Memory~\citep{DBLP:journals/corr/abs-2506-07398} \\
        PDDL~\citep{DBLP:conf/nips/MaZZYYJLKH24} & 20 & 40 & SR & G-Memory~\citep{DBLP:journals/corr/abs-2506-07398} \\
        BabyAI~\citep{DBLP:conf/nips/MaZZYYJLKH24} & 20 & 92 & SR & Evo-Memory~\citep{DBLP:journals/corr/abs-2511-20857} \\
        ScienceWorld~\citep{DBLP:conf/nips/MaZZYYJLKH24} & 20 & 70 & SR & Evo-Memory~\citep{DBLP:journals/corr/abs-2511-20857} \\
        ToolBench~\citep{DBLP:conf/acl/GuoCWLQLL0L24} & -- & 330 & SoPR & Evo-Memory~\citep{DBLP:journals/corr/abs-2511-20857} \\
        \bottomrule
    \end{tabular}}
    \caption{Datasets used in \benchmarkname{}, grouped by task category. $R$ denotes the reward function: Acc.\ = accuracy, EM = exact match, Rule = rule-based verification, SR = success rate, Pass@1 = code execution pass rate, SoPR = solvable pass rate (LLM judge), LLM Judge = LLM-as-judge evaluation. \textit{Memory Work} indicates the prior work that uses each dataset for memory evaluation.}
    \label{tab:benchmark}
\end{table}

\subsection{Self-Evolving Frameworks}
\label{app:self-evolve}

Since we focus on \textit{self}-evolution, all LLMs used for optimization are the same as $\mathrm{LLM}_e$. We describe the framework-specific hyperparameters below.

\paragraph{GEPA.} Following \citet{DBLP:journals/corr/abs-2507-19457}, we reserve 100 examples from the training set as a validation set and optimize over the remaining 230 examples. At each reflection step, the Proposer conditions on a minibatch of 5 training logs together with the current prompt to propose a refined candidate. Each candidate that improves on the minibatch is then evaluated on the validation set to determine whether it is added to the candidate pool.

\paragraph{ACE.} We adopt the online setting of \citet{DBLP:journals/corr/abs-2510-04618}, where ACE optimizes over all 330 training examples in a single epoch without a validation set. For each training example, the Reflector analyzes the extraction attempt and may iterate up to 3 reflection rounds when the attempt is incorrect. The Curator aggregates helpful/harmful signals and applies atomic operations (add, update, delete) to the playbook every 25 steps.

\paragraph{MemEvolve.} MemEvolve runs tournament-based evolution over \texttt{num\_rounds}${}=5$ rounds. In each round, the Analyzer (a tool-augmented agent) inspects the seed prompt's training logs on \texttt{batch\_x}${}=35$ examples, and the Proposer generates 3 new candidate prompts based on the analysis. All 4 prompts are then evaluated on \texttt{batch\_x}${}=35$ newly sampled examples plus \texttt{extra\_sample\_y}${}=10$ samples drawn from the current batch for stability, and the winner advances to the next round.

Our \methodname{} adopts the exact same hyperparameters as MemEvolve.

\subsection{Continual Memory Extraction}
\label{app:continual-details}

The \textit{No Memory} baselines in Table~\ref{tab:continual} differ between the single-step and continual settings for the following two reasons:

\paragraph{Different evaluated examples.} In the single-step setting, each evaluation instance is a triple (source conversation $c$, target query $q$, reward function $R$): $\mathrm{LLM}_e$ extracts memory from $c$, and $\mathrm{LLM}_g$ answers $q$ with or without that memory. In the continual setting, examples arrive as a sequential stream without paired source conversations; evaluation is performed on each incoming example using memories retrieved from previously seen examples. Because the two settings draw from different example pools, the \textit{No Memory} baselines naturally differ.

\paragraph{Different few-shot setup for AlfWorld.} Following G-Memory~\citep{DBLP:journals/corr/abs-2506-07398}, AlfWorld uses a 1-shot in-context demonstration for memory extraction. In the continual setting, every example undergoes extraction and thus includes this demonstration; in the single-step setting, only the source conversation requires extraction, so we remove the demonstration when answering the target query (0-shot) to better isolate the contribution of extracted memory. This difference accounts for the substantially higher \textit{No Memory} baseline in the continual setting.

Additionally, we note that the continual setting introduces a challenge not addressed by our current evolution pipeline: the extraction model must process source conversations that already contain retrieved memories, yet our prompts are evolved exclusively under the single-step regime where source conversations are memory-free. The experiments in \S\ref{sec:continual} serve to verify that a good prompt evolved in the single-step setting retains its advantages in the continual setting; we leave evolving prompts directly under the continual and heterogeneous regime to future work.

\section{Additional Experiments and Results}

\subsection{Generalizing Across Extraction Backends}
\label{app:eval-gemini}

We test whether \methodname{} generalizes across different extraction backends. Since our focus is on self-evolving frameworks, we treat the extraction model $\mathrm{LLM}_e$ and the optimization LLM as a single memory extraction system and use the same backend for both. We repeat the evolution pipeline with Gemini-3-Flash as the extraction backend, starting from the \textit{Simple} prompt.

\begin{table}[ht]
    \small
    \centering
    \resizebox{\linewidth}{!}{
    \begin{tabular}{lcccccccc}
        \toprule
         & \multicolumn{2}{c}{\textbf{Personalization}} & \multicolumn{2}{c}{\textbf{Problem-Solving}} & \multicolumn{2}{c}{\textbf{Agentic}} & \multicolumn{2}{c}{\textbf{Overall}} \\
         \cmidrule(lr){2-3}\cmidrule(lr){4-5}\cmidrule(lr){6-7}\cmidrule(lr){8-9}
         & \textbf{MA} & \textbf{RG} & \textbf{MA} & \textbf{RG} & \textbf{MA} & \textbf{RG} & \textbf{MA} & \textbf{RG} \\
         \midrule
         \textbf{No Memory} & 34.15 & -- & 46.52 & -- & 30.36 & -- & 37.84 & -- \\
         \textbf{Simple} & 65.86 & 0 & 50.19 & 0 & \hlfirst{38.46} & 0 & 50.46 & 0 \\
         \midrule
         \textbf{GEPA} & \hlsecond{68.59} & $+$5.64 & \hlsecond{52.91} & $+$7.71 & 33.77 & $-$9.99 & \hlsecond{50.71} & $+$0.93 \\
         \textbf{ACE} & 66.65 & $-$0.43 & 48.43 & $-$1.98 & 34.45 & $-$8.88 & 48.63 & $-$3.93 \\
         \textbf{MemEvolve} & 65.46 & $-$0.93 & 51.05 & $+$2.37 & \hlsecond{36.84} & $-$2.89 & 50.15 & $-$0.29 \\
         \textbf{\methodname{}} & \hlfirst{69.64} & \textbf{$+$6.66} & \hlfirst{53.64} & \textbf{$+$9.59} & 35.85 & $-$5.93 & \hlfirst{51.98} & \textbf{$+$3.40} \\
        \bottomrule
    \end{tabular}%
    }
    \caption{Evaluation of \textbf{Gemini-3-Flash} as the extraction backend from the \textit{Simple} prompt. \textbf{RG} denotes the Relative Gain (\%) over \textit{Simple}.}
    \label{tab:evaluation-gemini}
\end{table}

As shown in Table~\ref{tab:evaluation-gemini}, \methodname{} again achieves the highest overall relative gain ($+$3.40\%) and leads on both Personalization ($+$6.66\%) and Problem-Solving ($+$9.59\%), confirming that the cluster-based evolution strategy transfers effectively to a different model family. Notably, the stronger Gemini-3-Flash backend already produces a substantially higher \textit{Simple} baseline than Qwen3-32B (50.46 vs.\ 46.00 overall macro accuracy), leaving less headroom for any optimization method. Despite this, \methodname{} still delivers meaningful gains, whereas all three baselines either regress or show only marginal improvement overall: ACE drops to $-$3.93\%, MemEvolve to $-$0.29\%, and GEPA achieves only $+$0.93\%.

On the Agentic category, all evolved prompts, including ours, fall below the \textit{Simple} baseline. We attribute this to the limited headroom on agentic tasks with a stronger backend: the \textit{Simple} prompt already achieves high accuracy, leaving little room for evolution to improve and making over-specification more likely to hurt. Nonetheless, \methodname{} still substantially outperforms the No Memory baseline, confirming that the extracted memories remain useful despite the regression from the seed prompt.

\subsection{Efficiency Analysis}

We also compare the computational efficiency of each self-evolving framework along three dimensions: wall-clock time, optimization LLM calls, and evaluation calls.
We define \textbf{optimization LLM calls} as the number of LLM invocations used exclusively for prompt optimization (e.g., analysis, generation, reflection, curation).
\textbf{Evaluation calls} count the total number of target queries answered by $\mathrm{LLM}_g$ throughout evolution.

\begin{table}[ht]
    \centering
    \begin{tabular}{lccc}
        \toprule
         & \textbf{Wall Time} & \textbf{Optimization LLM Calls} & \textbf{Eval Calls} \\
        \midrule
        \textbf{GEPA}      & $\sim$7.4h  & 46   & 1,645 \\
        \textbf{ACE}       & $\sim$12.4h & 610  & 1,120 \\
        \textbf{MemEvolve} & $\sim$5.0h  & 30   & 1,150 \\
        \textbf{\methodname{}}      & $\sim$5.5h  & 221  & 1,150 \\
        \bottomrule
    \end{tabular}
    \caption{Efficiency comparison of self-evolving frameworks.}
    \label{tab:efficiency}
\end{table}

As noted by \citet{DBLP:journals/corr/abs-2507-19457}, the majority of GEPA's evaluation calls are consumed by validation-set scoring, which serves only for candidate selection without producing learning signals. This results in the highest evaluation cost (1,645 calls) and contributes to its longer wall time.
ACE, by contrast, has the fewest evaluation calls (1,120) since it forgoes maintaining a candidate pool entirely; however, its online learning loop is inherently sequential, as each step must complete extraction, reflection, and optional retry rounds before advancing, which severely limits parallelism and makes it the slowest method overall ($\sim$12.4h).
Our method builds on MemEvolve's tournament-based evolution by adding per-example summarization and cluster-based error analysis, introducing moderate optimizer overhead (221 vs.\ 30 calls) at a marginal wall-time cost ($\sim$5.5h vs.\ $\sim$5.0h). Overall, \methodname{} achieves the strongest performance gains (Table~\ref{tab:evaluation-simple}) with only a marginal increase in wall time over the most efficient baseline.

\subsection{Cluster Evolution and Taxonomy Construction}
\label{app:cluster-evolution}

Below we analyze how CluE evolves clusters starting from \textit{Simple} and using Qwen3-32B.

\paragraph{Cluster evolution.}
The cluster pool starts with 7 fine-grained clusters and consolidates over rounds as the system discovers that initially separate scenarios share common extraction patterns. For example, \emph{Emotional Context and Relational Dynamics} is absorbed into \emph{User Preferences}, since both require capturing user-specific states from conversational cues; similarly, \emph{Combinatorial and Puzzle Problem-Solving} merges into \emph{Technical Problem-Solving}, as both benefit from extracting multi-step reasoning chains. After the final round, the pool contains four clusters:
\begin{enumerate}
    \item User Preferences and Emotional Context (with Translation)
    \item Factual Data Disambiguation and Verification
    \item Procedural Knowledge in Virtual Environments
    \item Technical and Scientific Problem-Solving
\end{enumerate}

While this evolution path is dominated by merges, new clusters can also emerge. For instance, in a separate run starting from the \textit{Survey} prompt, the system splits out a \emph{Code-based Technical Workflows} cluster from the broader \emph{Technical Problem-Solving} cluster at Round~2, recognizing that code-related examples require extracting implementation-specific patterns (e.g., dependency management, error handling) distinct from general reasoning strategies.

\paragraph{From clusters to memory taxonomy.}
The proposer synthesizes per-cluster analyses into a unified extraction prompt, where taxonomy categories need not map one-to-one to clusters (Figure~\ref{fig:prompt_proposer}). In the final evolved prompt (Figure~\ref{fig:prompt_evolved}), the four clusters yield five taxonomy sections:
\begin{enumerate}
    \item Factual Data \& Temporal Disambiguation ($\leftarrow$ cluster~2)
    \item User Preferences \& Emotional Context ($\leftarrow$ cluster~1)
    \item Procedural \& Technical Knowledge ($\leftarrow$ cluster~3)
    \item Logical \& Combinatorial Reasoning ($\leftarrow$ cluster~4)
    \item Translation \& Stylistic Requirements ($\leftarrow$ cluster~1)
\end{enumerate}
Notably, cluster~1, which mixes user-preference extraction with translation-style tasks, is split into two taxonomy sections (2 and 5). This decoupled design allows the taxonomy to reorganize cluster-level insights into clean, non-overlapping categories without being constrained by cluster boundaries.

\subsection{Expanded Results}

Table~\ref{tab:per-dataset}, ~\ref{tab:per-dataset-seed}, ~\ref{tab:per-dataset-gemini} report the per-dataset accuracy for every method on the in-distribution test sets of \benchmarkname{}. As discussed in Section~\ref{sec:eval-evolving}, our goal is broad generalizability rather than peak performance on any single dataset. While individual dataset scores may fluctuate across methods, category-level and overall aggregates provide a more reliable picture of extraction quality, as they smooth out dataset-specific variance.

\begin{table}[ht]
    \small
    \centering
    \resizebox{\linewidth}{!}{
    \begin{tabular}{lllllll}
        \toprule
        \textbf{Dataset} & \textbf{No Memory} & \textbf{Simple} & \textbf{GEPA} & \textbf{ACE} & \textbf{MemEvolve} & \textbf{\methodname{}} \\
        \midrule
        \multicolumn{7}{c}{\textbf{\textit{Personalization}}} \\
        \midrule
        \textbf{MemBench-High}    & 20.50 & \hlfirst{87.50} & 83.00 & 77.00 & \hlsecond{87.00} & \hlsecond{87.00} \\
        \textbf{MemBench-Low}     & 55.00$_{\pm 1.77}$ & 60.83$_{\pm 1.18}$ & 59.58$_{\pm 1.56}$ & 58.33$_{\pm 3.12}$ & \hlfirst{64.58}$_{\pm 2.12}$ & \hlsecond{63.75}$_{\pm 1.02}$ \\
        \textbf{PersonaMem}       & 19.12 & 21.69 & 22.79 & \hlfirst{26.10} & \hlfirst{26.10} & \hlsecond{24.63} \\
        \textbf{PrefEval}         & 42.00 & 65.00 & 61.00 & 53.50 & \hlsecond{77.00} & \hlfirst{87.50} \\
        \midrule
        \multicolumn{7}{c}{\textbf{\textit{Problem-Solving}}} \\
        \midrule
        \textbf{GameOf24}         & 26.67$_{\pm 3.86}$ & 27.92$_{\pm 3.86}$ & 32.08$_{\pm 3.12}$ & \hlfirst{38.75}$_{\pm 7.14}$ & 27.50$_{\pm 2.70}$ & \hlsecond{37.08}$_{\pm 2.12}$ \\
        \textbf{AIME}             & 27.97$_{\pm 1.14}$ & 28.67$_{\pm 1.14}$ & 27.74$_{\pm 0.87}$ & \hlsecond{31.00}$_{\pm 0.33}$ & \hlfirst{31.47}$_{\pm 2.62}$ & 27.51$_{\pm 1.74}$ \\
        \textbf{MMLU-Pro Physics} & 70.50 & \hlfirst{73.00} & \hlsecond{72.50} & 71.00 & 72.00 & \hlfirst{73.00} \\
        \textbf{MMLU-Pro Eng.}    & 42.00 & 47.00 & \hlfirst{54.50} & \hlsecond{51.50} & \hlsecond{51.50} & 51.00 \\
        \textbf{MathEqBal}        & 87.00 & 93.00 & 92.00 & 90.50 & \hlsecond{93.50} & \hlfirst{96.50} \\
        \textbf{BigCodeBench}     & \hlsecond{25.00} & 23.00 & 21.50 & 23.50 & 21.00 & \hlfirst{26.00} \\
        \midrule
        \multicolumn{7}{c}{\textbf{\textit{Agentic}}} \\
        \midrule
        \textbf{ALFWorld}         & 30.41$_{\pm 2.71}$ & 47.66$_{\pm 2.19}$ & \hlfirst{57.02}$_{\pm 2.48}$ & 42.11$_{\pm 3.28}$ & 40.64$_{\pm 0.41}$ & \hlsecond{55.85}$_{\pm 3.23}$ \\
        \textbf{FEVER}            & 37.00 & 36.50 & \hlfirst{42.00} & \hlsecond{40.00} & 35.50 & \hlsecond{40.00} \\
        \textbf{PDDL}             & \hlsecond{18.33}$_{\pm 1.18}$ & 16.67$_{\pm 2.36}$ & \hlfirst{19.17}$_{\pm 1.18}$ & \hlsecond{18.33}$_{\pm 3.12}$ & 15.00 & 16.67$_{\pm 3.12}$ \\
        \textbf{BabyAI}           & \hlfirst{32.25}$_{\pm 1.02}$ & 21.01$_{\pm 3.36}$ & 23.19$_{\pm 1.36}$ & 23.19$_{\pm 3.12}$ & \hlsecond{25.36}$_{\pm 2.23}$ & 23.19$_{\pm 1.36}$ \\
        \textbf{ScienceWorld}     & 33.81$_{\pm 3.37}$ & 40.48$_{\pm 2.94}$ & \hlfirst{44.76}$_{\pm 3.75}$ & \hlsecond{43.81}$_{\pm 1.78}$ & 38.10$_{\pm 1.35}$ & 40.48$_{\pm 4.71}$ \\
        \bottomrule
    \end{tabular}}
    \caption{Per-dataset accuracy (\%) on in-distribution test sets for Qwen3-32B evolving from \textit{Simple}. Subscripts denote standard deviations over three runs.}
    \label{tab:per-dataset}
\end{table}

\begin{table}[ht]
    \small
    \centering
    \resizebox{\linewidth}{!}{
    \begin{tabular}{lllllll}
        \toprule
        \textbf{Dataset} & \textbf{No Memory} & \textbf{Survey} & \textbf{GEPA*} & \textbf{ACE} & \textbf{MemEvolve} & \textbf{\methodname{}} \\
        \midrule
        \multicolumn{7}{c}{\textbf{\textit{Personalization}}} \\
        \midrule
        \textbf{MemBench-High}    & 20.50 & \hlfirst{87.00} & \hlfirst{87.00} & 82.00 & 85.50 & \hlsecond{86.00} \\
        \textbf{MemBench-Low}     & 55.00$_{\pm 1.77}$ & 64.58$_{\pm 6.24}$ & 64.58$_{\pm 6.24}$ & 57.08$_{\pm 1.18}$ & \hlsecond{67.08}$_{\pm 2.36}$ & \hlfirst{80.83}$_{\pm 3.12}$ \\
        \textbf{PersonaMem}       & 19.12 & \hlsecond{24.63} & \hlsecond{24.63} & \hlsecond{24.63} & 22.43 & \hlfirst{26.84} \\
        \textbf{PrefEval}         & 42.00 & 72.00 & 72.00 & 55.50 & \hlsecond{76.50} & \hlfirst{89.00} \\
        \midrule
        \multicolumn{7}{c}{\textbf{\textit{Problem-Solving}}} \\
        \midrule
        \textbf{GameOf24}         & 26.67$_{\pm 3.86}$ & \hlsecond{34.17}$_{\pm 6.80}$ & \hlsecond{34.17}$_{\pm 6.80}$ & 32.50$_{\pm 1.02}$ & \hlfirst{34.58}$_{\pm 1.56}$ & \hlfirst{34.58}$_{\pm 2.57}$ \\
        \textbf{AIME}             & 27.97$_{\pm 1.14}$ & 31.00$_{\pm 2.64}$ & 31.00$_{\pm 2.64}$ & 29.84$_{\pm 1.44}$ & \hlsecond{31.24}$_{\pm 2.16}$ & \hlfirst{31.70}$_{\pm 2.01}$ \\
        \textbf{MMLU-Pro Physics} & 70.50 & 71.00 & 71.00 & \hlsecond{71.50} & \hlfirst{72.00} & 70.50 \\
        \textbf{MMLU-Pro Eng.}    & 42.00 & 50.00 & 50.00 & 47.00 & \hlfirst{55.50} & \hlsecond{51.00} \\
        \textbf{MathEqBal}        & 87.00 & 90.00 & 90.00 & \hlsecond{92.50} & 91.00 & \hlfirst{98.50} \\
        \textbf{BigCodeBench}     & \hlsecond{25.00} & 23.00 & 23.00 & \hlfirst{27.00} & 23.00 & 23.50 \\
        \midrule
        \multicolumn{7}{c}{\textbf{\textit{Agentic}}} \\
        \midrule
        \textbf{ALFWorld}         & 30.41$_{\pm 2.71}$ & \hlsecond{47.37}$_{\pm 3.12}$ & \hlsecond{47.37}$_{\pm 3.12}$ & 46.49$_{\pm 3.28}$ & 40.35$_{\pm 3.12}$ & \hlfirst{49.42}$_{\pm 4.07}$ \\
        \textbf{FEVER}            & 37.00 & \hlsecond{40.00} & \hlsecond{40.00} & \hlfirst{42.00} & 36.00 & \hlsecond{40.00} \\
        \textbf{PDDL}             & \hlsecond{18.33}$_{\pm 1.18}$ & 16.67$_{\pm 3.12}$ & 16.67$_{\pm 3.12}$ & 16.67$_{\pm 1.18}$ & 15.83$_{\pm 2.36}$ & \hlfirst{20.00}$_{\pm 2.04}$ \\
        \textbf{BabyAI}           & \hlfirst{32.25}$_{\pm 1.02}$ & 21.01$_{\pm 2.71}$ & 21.01$_{\pm 2.71}$ & \hlsecond{25.00}$_{\pm 2.66}$ & 23.55$_{\pm 2.56}$ & 23.55$_{\pm 1.02}$ \\
        \textbf{ScienceWorld}     & 33.81$_{\pm 3.37}$ & \hlfirst{42.86}$_{\pm 4.21}$ & \hlfirst{42.86}$_{\pm 4.21}$ & \hlsecond{41.90}$_{\pm 2.94}$ & 40.95$_{\pm 4.71}$ & 40.48$_{\pm 4.10}$ \\
        \bottomrule
    \end{tabular}}
    \caption{Per-dataset accuracy (\%) on in-distribution test sets for Qwen3-32B evolving from \textit{Survey}.}
    \label{tab:per-dataset-seed}
\end{table}

\begin{table}[ht]
    \small
    \centering
    \resizebox{\linewidth}{!}{
    \begin{tabular}{lllllll}
        \toprule
        \textbf{Dataset} & \textbf{No Memory} & \textbf{Simple} & \textbf{GEPA} & \textbf{ACE} & \textbf{MemEvolve} & \textbf{\methodname{}} \\
        \midrule
        \multicolumn{7}{c}{\textbf{\textit{Personalization}}} \\
        \midrule
        \textbf{MemBench-High}    & 20.50 & 84.50 & 84.00 & \hlsecond{86.50} & 84.50 & \hlfirst{87.00} \\
        \textbf{MemBench-Low}     & 55.00$_{\pm 1.77}$ & 52.50$_{\pm 4.45}$ & \hlsecond{60.00}$_{\pm 1.02}$ & 57.08$_{\pm 1.56}$ & 52.50$_{\pm 2.70}$ & \hlfirst{60.42}$_{\pm 2.57}$ \\
        \textbf{PersonaMem}       & 19.12 & 34.93 & \hlfirst{37.87} & 30.51 & 33.82 & \hlsecond{37.13} \\
        \textbf{PrefEval}         & 42.00 & 91.50 & \hlsecond{92.50} & \hlsecond{92.50} & 91.00 & \hlfirst{94.00} \\
        \midrule
        \multicolumn{7}{c}{\textbf{\textit{Problem-Solving}}} \\
        \midrule
        \textbf{GameOf24}         & 26.67$_{\pm 3.86}$ & 35.42$_{\pm 7.66}$ & \hlsecond{45.00}$_{\pm 1.77}$ & 31.67$_{\pm 1.18}$ & 36.67$_{\pm 5.80}$ & \hlfirst{50.42}$_{\pm 2.12}$ \\
        \textbf{AIME}             & 27.97$_{\pm 1.14}$ & 31.70$_{\pm 0.66}$ & 31.47$_{\pm 1.51}$ & 28.44$_{\pm 0.66}$ & \hlfirst{32.63}$_{\pm 1.44}$ & \hlsecond{31.93}$_{\pm 1.65}$ \\
        \textbf{MMLU-Pro Physics} & 70.50 & \hlsecond{74.00} & \hlsecond{74.00} & 72.00 & \hlfirst{74.50} & 72.50 \\
        \textbf{MMLU-Pro Eng.}    & 42.00 & 44.50 & \hlfirst{51.50} & \hlsecond{51.00} & 48.50 & \hlsecond{51.00} \\
        \textbf{MathEqBal}        & 87.00 & \hlfirst{94.00} & 92.00 & 83.50 & \hlsecond{92.50} & \hlsecond{92.50} \\
        \textbf{BigCodeBench}     & \hlfirst{25.00} & 21.50 & 23.50 & \hlsecond{24.00} & 21.50 & 23.50 \\
        \midrule
        \multicolumn{7}{c}{\textbf{\textit{Agentic}}} \\
        \midrule
        \textbf{ALFWorld}         & 30.41$_{\pm 2.71}$ & \hlfirst{54.97}$_{\pm 1.80}$ & 40.06$_{\pm 2.98}$ & 42.98$_{\pm 2.48}$ & \hlsecond{51.75}$_{\pm 3.12}$ & 51.46$_{\pm 3.68}$ \\
        \textbf{FEVER}            & 37.00 & \hlfirst{45.00} & 36.00 & \hlsecond{40.50} & \hlsecond{40.50} & \hlsecond{40.50} \\
        \textbf{PDDL}             & \hlsecond{18.33}$_{\pm 1.18}$ & 16.67$_{\pm 2.36}$ & \hlsecond{18.33}$_{\pm 3.12}$ & 15.83$_{\pm 4.25}$ & \hlfirst{20.00}$_{\pm 2.04}$ & \hlfirst{20.00}$_{\pm 2.04}$ \\
        \textbf{BabyAI}           & \hlfirst{32.25}$_{\pm 1.02}$ & \hlsecond{22.83}$_{\pm 3.55}$ & 20.65$_{\pm 1.77}$ & 22.46$_{\pm 0.51}$ & 19.57$_{\pm 0.89}$ & 17.75$_{\pm 4.47}$ \\
        \textbf{ScienceWorld}     & 33.81$_{\pm 3.37}$ & \hlsecond{52.86}$_{\pm 5.35}$ & \hlfirst{53.81}$_{\pm 3.75}$ & 50.48$_{\pm 1.35}$ & 52.38$_{\pm 2.43}$ & 49.52$_{\pm 4.86}$ \\
        \bottomrule
    \end{tabular}}
    \caption{Per-dataset accuracy (\%) on in-distribution test sets for Gemini-3-Flash evolving from \textit{Simple}.}
    \label{tab:per-dataset-gemini}
\end{table}

\section{Prompts}

\subsection{Static Memory Extraction Prompts}
\label{app:static-prompts}

We describe the five static extraction prompts evaluated in Table~\ref{tab:evaluation-static}. \textit{Simple} (Figure~\ref{fig:prompt_simple}) is a minimal baseline that instructs the model to extract useful information with an empty taxonomy and minimal guidelines. \textit{Mem0} and \textit{ReasoningBank} provide detailed, category-specific extraction guidelines: \textit{Mem0} (Figure~\ref{fig:prompt_mem0}) targets personalization-oriented facts such as preferences, relationships, and plans, while \textit{ReasoningBank} (Figure~\ref{fig:prompt_reasoningbank}) extracts strategic insights and lessons from successful and failed trajectories. In contrast, \textit{OpenMemory} and \textit{Survey} define broader memory taxonomies: \textit{OpenMemory} (Figure~\ref{fig:prompt_openmemory}) uses a five-class taxonomy (episodic, semantic, procedural, emotional, reflective), while \textit{Survey} (Figure~\ref{fig:prompt_survey}) uses a two-class taxonomy (factual and experiential). 

\subsection{Prompts for \methodname{}}
\label{app:clue-prompt}

The prompts for the different LLM roles in the \methodname{} optimization pipeline are provided in Figures~\ref{fig:prompt_summarizer}, \ref{fig:prompt_cluster_manager}, \ref{fig:prompt_cluster_analyzer}, and~\ref{fig:prompt_proposer}.

\begin{center} % 仅用于整体居中（可选）
\begin{tcolorbox}[
  enhanced,
  breakable,                 % 允许跨页
  colback=white,
  colframe=black!50!white,
  title={\textit{Simple} prompt},
  width=1.0\linewidth        % 控制盒子宽度
]
\lstset{style=boxedcode}
\begin{lstlisting}
# Role
You are an expert **Memory Extraction Agent**. Your goal is to extract reusable, high-value memories from a single conversation, in order to benefit the assistant in future interactions where this conversation is no longer accessible.

# Input Format
The conversation is wrapped inside `<conversation>` tags, containing:
- `<system>`: System instructions.
- `<user>`: User messages.
- `<assistant>`: Assistant responses.

# Memory Taxonomy

# General Guidelines
- Extract at most 5 memories in total.

# Output Format
Output only the memories as a numbered list.
1. Memory 1
2. Memory 2
...
\end{lstlisting}
\end{tcolorbox}
\captionsetup{justification=centering} % 可选：让标题居中
\captionof{figure}{The \textit{Simple} prompt.}
\label{fig:prompt_simple}
\end{center}

\begin{center}
\begin{tcolorbox}[
  enhanced,
  breakable,
  colback=white,
  colframe=black!50!white,
  title={\textit{Mem0} prompt},
  width=1.0\linewidth
]
\lstset{style=boxedcode}
\begin{lstlisting}
You are an Memory Extraction Agent, specialized in accurately storing facts, user memories, and preferences. Your primary role is to extract relevant pieces of information from conversations and organize them into distinct, manageable facts. This allows for easy retrieval and personalization in future interactions. Below are the types of information you need to focus on and the detailed instructions on how to handle the input data.

Types of Information to Remember:

1. Store Personal Preferences: Keep track of likes, dislikes, and specific preferences in various categories such as food, products, activities, and entertainment.
2. Maintain Important Personal Details: Remember significant personal information like names, relationships, and important dates.
3. Track Plans and Intentions: Note upcoming events, trips, goals, and any plans the user has shared.
4. Remember Activity and Service Preferences: Recall preferences for dining, travel, hobbies, and other services.
5. Monitor Health and Wellness Preferences: Keep a record of dietary restrictions, fitness routines, and other wellness-related information.
6. Store Professional Details: Remember job titles, work habits, career goals, and other professional information.
7. Miscellaneous Information Management: Keep track of favorite books, movies, brands, and other miscellaneous details that the user shares.

Here are some few shot examples:

Input: Hi.
Output: {"facts" : []}

Input: There are branches in trees.
Output: {"facts" : []}

Input: Hi, I am looking for a restaurant in San Francisco.
Output: {"facts" : ["Looking for a restaurant in San Francisco"]}

Input: Yesterday, I had a meeting with John at 3pm. We discussed the new project.
Output: {"facts" : ["Had a meeting with John at 3pm", "Discussed the new project"]}

Input: Hi, my name is John. I am a software engineer.
Output: {"facts" : ["Name is John", "Is a Software engineer"]}

Input: Me favourite movies are Inception and Interstellar.
Output: {"facts" : ["Favourite movies are Inception and Interstellar"]}

Return the facts and preferences in a json format as shown above.

Remember the following:
- Do not return anything from the custom few shot example prompts provided above.
- If you do not find anything relevant in the below conversation, you can return an empty list corresponding to the "facts" key.
- Create the facts based on the user and assistant messages only. Do not pick anything from the system messages.
- Make sure to return the response in the format mentioned in the examples. The response should be in json with a key as "facts" and corresponding value will be a list of strings.

Given a conversation between the user and the assistant, you have to extract the relevant facts and preferences about the user, if any, from the conversation and return them in the json format as shown above.
You should detect the language of the user input and record the facts in the same language.

# Input Format
The conversation is wrapped inside `<conversation>` tags, containing:
- `<system>`: System instructions.
- `<user>`: User messages.
- `<assistant>`: Assistant responses.

# Output Format
Output only the facts in a JSON format as shown in the examples above.
\end{lstlisting}
\end{tcolorbox}
\captionsetup{justification=centering}
\captionof{figure}{The \textit{Mem0} prompt.}
\label{fig:prompt_mem0}
\end{center}

\begin{center}
\begin{tcolorbox}[
  enhanced,
  breakable,
  colback=white,
  colframe=black!50!white,
  title={\textit{ReasoningBank} prompt},
  width=1.0\linewidth
]
\lstset{style=boxedcode}
\begin{lstlisting}
# Role
You are an expert **Memory Extraction Agent**. Your goal is to extract reusable, high-value memories from a single conversation, in order to benefit the assistant in future interactions where this conversation is no longer accessible.

# Input Format
The conversation is wrapped inside `<conversation>` tags, containing:
- `<system>`: System instructions.
- `<user>`: User messages.
- `<assistant>`: Assistant responses.

# Guidelines
You need to extract and summarize useful insights in the format of memory items based on the conversation.
The goal of summarized memory items is to be helpful and generalizable for future similar tasks.

# Important notes
- For successful conversations, you must first think why the conversation was successful, and then summarize the insights.
- For failed conversations, you must first reflect and think why the conversation failed, and then summarize what lessons you have learned or strategies to prevent the failure in the future.
- You can extract at most 3 memory items from the conversation.
- You must not repeat similar or overlapping items.
- Do not mention specific websites, queries, or string contents, but rather focus on the generalizable insights.

# Output Format
Output only the memories as a numbered list.
1. Memory 1
2. Memory 2
3. Memory 3
\end{lstlisting}
\end{tcolorbox}
\captionsetup{justification=centering}
\captionof{figure}{The \textit{ReasoningBank} prompt.}
\label{fig:prompt_reasoningbank}
\end{center}

\begin{center}
\begin{tcolorbox}[
  enhanced,
  breakable,
  colback=white,
  colframe=black!50!white,
  title={\textit{OpenMemory} prompt},
  width=1.0\linewidth
]
\lstset{style=boxedcode}
\begin{lstlisting}
# Role
You are an expert **Memory Extraction Agent**. Your goal is to extract reusable, high-value memories from a single conversation, in order to benefit the assistant in future interactions where this conversation is no longer accessible.

# Input Format
The conversation is wrapped inside `<conversation>` tags, containing:
- `<system>`: System instructions.
- `<user>`: User messages.
- `<assistant>`: Assistant responses.

# Memory Taxonomy
## Episodic Memory
- Time-bound events & experiences, e.g., "yesterday I attended a conference".

## Semantic Memory
- Timeless facts & knowledge, e.g., "Paris is the capital of France".

## Procedural Memory
- Skills, procedures, how-tos, e.g. "to deploy: build, test, push".

## Emotional Memory
- Feelings, sentiment, mood, e.g. "I'm excited about this project!"

## Reflective Memory
- Meta-cognition, insights, e.g. "I learn best through practice".

# General Guidelines
- Extract at most 5 memories in total.

# Output Format
Output only the memories as a numbered list.
1. Memory 1
2. Memory 2
...
\end{lstlisting}
\end{tcolorbox}
\captionsetup{justification=centering}
\captionof{figure}{The \textit{OpenMemory} prompt.}
\label{fig:prompt_openmemory}
\end{center}

\begin{center}
\begin{tcolorbox}[
  enhanced,
  breakable,
  colback=white,
  colframe=black!50!white,
  title={\textit{Survey} prompt},
  width=1.0\linewidth
]
\lstset{style=boxedcode}
\begin{lstlisting}
# Role
You are an expert **Memory Extraction Agent**. Your goal is to extract reusable, high-value memories from a single conversation, in order to benefit the assistant in future interactions where this conversation is no longer accessible.

# Input Format
The conversation is wrapped inside `<conversation>` tags, containing:
- `<system>`: System instructions.
- `<user>`: User messages.
- `<assistant>`: Assistant responses.

# Memory Taxonomy
## Factual Memory
Factual memory refers to the assistant's declarative knowledge base, established to ensure consistency, coherence, and adaptability by recalling explicit facts, user preferences, and environmental states.

## Experiential Memory
Experiential memory refers to the assistant's procedural and strategic knowledge, accumulated to enable continual learning and self-evolution by abstracting from past trajectories, failures, and successes.

# General Guidelines
- Extract at most 5 memories in total.

# Output Format
Output only the memories as a numbered list.
1. Memory 1
2. Memory 2
...
\end{lstlisting}
\end{tcolorbox}
\captionsetup{justification=centering}
\captionof{figure}{The \textit{Survey} prompt.}
\label{fig:prompt_survey}
\end{center}

\begin{center}
\begin{tcolorbox}[
  enhanced,
  breakable,
  colback=white,
  colframe=black!50!white,
  title={Prompt for Summarizer},
  width=1.0\linewidth
]
\lstset{style=boxedcode}
\begin{lstlisting}
You are analyzing a memory extraction example to produce an extraction-targeted summary.

Your summary should focus on the EXTRACTION SCENARIO -- NOT on the surface-level topic or dataset.

Given the following extraction result, write a concise 2-3 sentence summary that describes:
1. What type of information needs to be extracted (e.g., factual knowledge, procedural steps, user preferences, causal reasoning, spatial/temporal relations, emotional context, etc.)
2. What makes this extraction challenging (e.g., long context, implicit information, multi-turn reasoning, noise/irrelevant content, ambiguous boundaries, need for abstraction vs verbatim capture, etc.)

source_conversation (first 4096 chars):
{source_preview}

extracted_memory:
{extracted_memory}

target_conversation (first 4096 chars):
{target_preview}

target_reward: {target_reward}

Write ONLY the extraction-targeted summary. No extra formatting, no bullet points -- just 2-3 flowing sentences.
\end{lstlisting}
\end{tcolorbox}
\captionsetup{justification=centering}
\captionof{figure}{Prompt for Summarizer.}
\label{fig:prompt_summarizer}
\end{center}

\begin{center}
\begin{tcolorbox}[
  enhanced,
  breakable,
  colback=white,
  colframe=black!50!white,
  title={Prompt for Cluster Manager},
  width=1.0\linewidth
]
\lstset{style=boxedcode}
\begin{lstlisting}
You are a clustering agent for memory extraction prompt evolution.

Your task is to group the training examples below into clusters based on EXTRACTION SCENARIO PATTERNS -- that is, what type of information needs to be extracted and what makes the extraction challenging. Do NOT cluster by surface-level task type, dataset name, or domain topic.

Good cluster labels describe extraction scenarios, for example:
- "Procedural knowledge in lengthy dialogue"
- "Implicit user preferences in casual multi-turn conversation"
- "Causal reasoning chains in problem-solving episodes"
- "Factual knowledge in noisy conversational context"
- "High-level strategies in step-by-step execution traces"

You may create up to 7 clusters. Use as many as naturally emerge from the data -- you do not need to reach the maximum. Each cluster must have at least 2 examples. If a pattern has only 1 example, merge it into the closest cluster.

{existing_pool_section}

Here are the extraction-targeted summaries for the current batch:
<batch_summaries>
{batch_summaries}
</batch_summaries>

Instructions:
- Assign EVERY example to exactly one cluster.
- If an existing cluster fits well, assign to it. You may update its label or description if the new examples reveal a better characterization.
- Create new clusters only when existing ones do not adequately capture the extraction scenario.
- You may merge two existing clusters if they are too similar, or split one if it covers clearly distinct patterns.
- Prefer stable clusters -- only change cluster definitions when there is strong evidence from the current batch.

Output format -- return ONLY a JSON object (no markdown, no explanation outside JSON):
{
  "clusters": [
    {
      "cluster_id": "cluster_XX",
      "label": "<short extraction scenario label>",
      "description": "<1-2 sentence description of what extraction scenario this cluster represents>",
      "example_task_ids": ["0000", "0003", ...]
    },
    ...
  ]
}
\end{lstlisting}
\end{tcolorbox}
\captionsetup{justification=centering}
\captionof{figure}{Prompt for Cluster Manager.}
\label{fig:prompt_cluster_manager}
\end{center}

\begin{center}
\begin{tcolorbox}[
  enhanced,
  breakable,
  colback=white,
  colframe=black!50!white,
  title={Prompt for Cluster Analyzer},
  width=1.0\linewidth
]
\lstset{style=boxedcode}
\begin{lstlisting}
You are an expert analysis agent for memory-extraction prompt evolution.
You are analyzing a SPECIFIC CLUSTER of examples that share a common extraction scenario.

Context:
- Round: {round_id}
- Cluster: {cluster_label}
- Cluster description: {cluster_description}
- Task IDs in this cluster: {task_ids}
- Total examples in cluster: {num_cluster_examples}
- Pair-level logs directory: {logs_dir}

Current base system prompt:
<base_system_prompt>
{base_prompt}
</base_system_prompt>

Your task:
1) Use tools to inspect extraction pair logs for the task IDs listed above ONLY.
   Focus exclusively on these tasks -- do not inspect tasks outside this cluster.
2) Analyze success and failure patterns SPECIFIC to this cluster's extraction scenario.
3) Propose prompt-level improvements targeted at this cluster's extraction scenario.

Hard constraints:
- Do not propose architecture changes.
- Do not propose online memory update loops.
- Focus only on improving extraction system prompt quality.
- Your analysis must be specific to this cluster's extraction scenario, not generic.

Required output:
Return a concise but structured report with sections:

**Success Analysis** (for high-reward examples in this cluster):
- Does the extracted memory assist the target request, and how does it help?
- What characteristics make this memory useful (e.g., level of abstraction, specificity, type of information, transferability)?
- What generalizable insights can be drawn for improving future memory extraction guidelines?

**Failure Analysis** (for low-reward examples in this cluster):
- What kind of memory extracted from the input could have better supported completing the target request?
- What was missing or ineffective in the extracted memory (e.g., missing key information, too specific, too vague, not transferable)?
- Based on this, how should the memory extraction guidelines be revised?

**Targeted Prompt Recommendations** (3-5 bullets specific to this cluster's scenario)
\end{lstlisting}
\end{tcolorbox}
\captionsetup{justification=centering}
\captionof{figure}{Prompt for Cluster Analyzer.}
\label{fig:prompt_cluster_analyzer}
\end{center}

\begin{center}
\begin{tcolorbox}[
  enhanced,
  breakable,
  colback=white,
  colframe=black!50!white,
  title={Prompt for Proposer},
  width=1.0\linewidth
]
\lstset{style=boxedcode}
\begin{lstlisting}
You are generating an improved memory extraction system prompt based on per-cluster analysis reports.

You have received analysis reports from {num_clusters} distinct extraction scenario clusters.
Your goal is to synthesize these cluster-specific insights into ONE improved system prompt that can handle diverse extraction scenarios effectively.

Cluster analyses:
<cluster_analyses>
{cluster_analyses_text}
</cluster_analyses>

Parent prompt id: {parent_prompt_id}
Base system prompt:
<base_system_prompt>
{base_system_prompt}
</base_system_prompt>

Synthesis instructions:
1. General guidelines: Identify recommendations that appear across multiple clusters. Distill these into clear, actionable general extraction guidelines that apply regardless of memory type.
2. Memory taxonomy: Group cluster-specific insights into memory categories. For each category, provide a concise definition and category-specific extraction guidelines. Categories do not need to map 1:1 to clusters -- merge or reorganize as needed to create a clean, non-overlapping taxonomy.
3. Conflict resolution: When cluster recommendations conflict, organize them under the appropriate memory category rather than forcing a single uniform rule. Different memory types may require different extraction approaches.
4. Stability: Make targeted, high-impact changes. Do not rewrite the entire prompt if focused additions or revisions suffice.

Prompt quality requirements -- the generated system prompt must be:
- Written in clear, direct language that another language model can follow without ambiguity.
- Structured with explicit sections and formatting so the extraction model can quickly locate relevant instructions.
- Actionable: use concrete action verbs (e.g., "extract", "summarize", "preserve", "omit") instead of vague descriptions (e.g., "be aware of", "consider").
- Include brief examples where they help clarify expected behavior, but keep them concise.

Constraints:
- Return JSON only (no markdown, no explanations outside JSON).
- Output must be a JSON array with exactly 1 object.
- Each object schema:
  {
    "candidate_id": "cand_XX",
    "system_prompt": "<full prompt text>",
    "rationale": "<short rationale explaining which cross-cluster insights drove the changes>"
  }
- Do not return empty prompts.
\end{lstlisting}
\end{tcolorbox}
\captionsetup{justification=centering}
\captionof{figure}{Prompt for Proposer.}
\label{fig:prompt_proposer}
\end{center}

\begin{center}
\begin{tcolorbox}[
  enhanced,
  breakable,
  colback=white,
  colframe=black!50!white,
  title={Evolved prompt},
  width=1.0\linewidth
]
\lstset{style=boxedcode}
\begin{lstlisting}
# Role
You are an expert **Memory Extraction Agent**. Your goal is to extract reusable, high-value memories from a single conversation to benefit future interactions.

# Input Format
The conversation is wrapped in `<conversation>` tags containing:
- `<system>`: System instructions
- `<user>`: User messages
- `<assistant>`: Assistant responses

# Memory Taxonomy

## 1. Factual Data & Temporal Disambiguation
- **Definition**: Memories resolving ambiguous entities, documenting verification attempts, or capturing verifiable facts with temporal/event anchors.
- **Guidelines**:
  - Append clarifiers like [film], [person] to ambiguous terms
  - Document all disambiguation attempts (e.g., search queries, NOT ENOUGH INFO conclusions)
  - For temporal conflicts, document both timestamps with message references and resolve contradictions using context
  - For verification claims, **explicitly document final classification (SUPPORTS/REFUTES/NOT ENOUGH INFO)**
  - Cross-verify all extracted factual claims against source conversation with 2+ message references
  - When primary entity search fails, document at least two alternative search paths with full result citations

## 2. User Preferences & Emotional Context
- **Definition**: Memories capturing categorical preferences, emotional states, or relational dynamics with temporal/event anchors.
- **Guidelines**:
  - Map each preference/emotion to **2+ distinct conversation threads** (not just messages)
  - Reject memories lacking 2+ distinct temporal anchors
  - Filter out transient emotional states mentioned <2 times
  - For translation/refinement tasks, extract only linguistic markers *explicitly* demonstrated in source examples
  - When encountering 'forget' commands, document both original and new states with timestamps
  - Include cross-topic consistency check: verify extracted preferences/emotions appear in at least 2 conversation threads

## 3. Procedural & Technical Knowledge
- **Definition**: Memories encoding command syntax, formulas, or implementation constraints.
- **Guidelines**:
  - Quote command syntax verbatim with at least 2 example usages from different contexts
  - For command extractions, **cross-check syntax against system-provided examples** and reject non-existent actions
  - Document sequential state transitions and error-handling patterns
  - For object manipulation tasks, explicitly list all preconditions (e.g., 'must be in same room')
  - For multi-step workflows, include recovery steps if intermediate actions fail
  - For technical components (e.g., skateboard rig), document only what is explicitly described in the source

## 4. Logical & Combinatorial Reasoning
- **Definition**: Memories abstracting logical rules, dimensionality-reduction strategies, or combinatorial formulas with validation.
- **Guidelines**:
  - **Cross-verify all extracted formulas/code against source conversation line-by-line**
  - Include at least 3 explicit constraints (input validation, edge cases, output requirements)
  - For physics problems, include sign convention validation and unit consistency checks
  - For multi-step calculations, include at least 2 distinct validation methods
  - Distinguish valid reasoning pathways from dead-ends using markers like [Valid] and [Exploratory]

## 5. Translation & Stylistic Requirements
- **Definition**: Memories capturing linguistic style preferences and translation constraints.
- **Guidelines**:
  - Extract only stylistic requirements (tone, formality) explicitly mentioned in source conversation
  - When requested stylistic adaptations, extract specific linguistic markers from source examples
  - For dynamic style changes (e.g., 'forget' commands), document explicit before/after state comparisons

# General Guidelines
- Extract at most 5 memories in total
- **Reject memories containing information not directly verifiable from source conversation**
- For multi-turn dialogues, abstract recurring patterns before specific examples
- For technical problems, distinguish rounding/truncation and handle multimodal edge cases
- Always cite conversation turns supporting each memory (e.g., 'User message 3: ...')
- **Cross-verify all extracted formulas, search result references, and message citations against source conversation**
- When extracting code logic, explicitly note deterministic requirements (e.g., random seed constraints)
- Require 2+ distinct temporal/event anchors for all preference/emotion memories
- For physics problems, include sign convention validation and unit consistency checks
- Map extracted memories to the 3-phase problem-solving framework when applicable

# Output Format
Output only the memories as a numbered list.
1. Memory 1
2. Memory 2
...
\end{lstlisting}
\end{tcolorbox}
\captionsetup{justification=centering}
\captionof{figure}{Prompt evolved by \methodname{} from \textit{Simple}, using Qwen3-32B.}
\label{fig:prompt_evolved}
\end{center}

\end{document}